\documentclass[journal]{IEEEtran}
\usepackage{cite}

\ifCLASSINFOpdf
   \usepackage[pdftex]{graphicx}
\else
\fi
\usepackage{amsmath,amssymb,amsfonts}
\ifCLASSOPTIONcompsoc
  \usepackage[caption=false,font=normalsize,labelfont=sf,textfont=sf]{subfig}
\else
  \usepackage[caption=false,font=footnotesize]{subfig}
\fi
\usepackage{url}

\usepackage{tikz}
\usepackage{hyperref}
\usepackage{multirow}
\usepackage{siunitx}
\usepackage{tabularx, booktabs}
\usepackage{xcolor}
\usepackage{enumitem}
\usepackage{textcomp}

\newcommand\copyrighttext{%
  \footnotesize \textcopyright 2022 IEEE. Personal use of this material is permitted.
  Permission from IEEE must be obtained for all other uses, in any current or future
  media, including reprinting/republishing this material for advertising or promotional
  purposes, creating new collective works, for resale or redistribution to servers or
  lists, or reuse of any copyrighted component of this work in other works.
  DOI: \href{https://doi.org/10.1109/TITS.2022.3155228}{10.1109/TITS.2022.3155228}}
\newcommand\copyrightnotice{%
\begin{tikzpicture}[remember picture,overlay]
\node[anchor=south,yshift=2pt] at (current page.south) {\fbox{\parbox{\dimexpr\textwidth-\fboxsep-\fboxrule\relax}{\copyrighttext}}};
\end{tikzpicture}%
}

\begin{document}
%
\title{Automatic Extrinsic Calibration Method for LiDAR and Camera Sensor Setups}
%
%
%

\author{Jorge~Beltr\'{a}n,
        Carlos~Guindel,
        Arturo de la Escalera,
        and~Fernando~Garc\'{i}a,~\IEEEmembership{Member,~IEEE}
\thanks{Manuscript submitted February 25, 2022. This work has been supported by the Madrid Government (Comunidad de Madrid) under the Multiannual Agreement with UC3M in the line of "Fostering Young Doctors Research" (PEAVAUTO-CM-UC3M) within the V PRICIT (5th Regional Programme of Research and Technological Innovation) and through SEGVAUTO-4.0-CM P2018/EMT-4362, and by the Spanish Government (RTI2018-096036-B-C21).}
\thanks{The authors are with the Department of Systems Engineering and Automation, Universidad Carlos III de Madrid, Leganés, 28911 Spain e-mail: \{jbeltran, cguindel, escalera, fegarcia\}@ing.uc3m.es).} 
}

\maketitle
\copyrightnotice

\begin{abstract}
Most sensor setups for onboard autonomous perception are composed of LiDARs and vision systems, as they provide complementary information that improves the reliability of the different algorithms necessary to obtain a robust scene understanding. However, the effective use of information from different sources requires an accurate calibration between the sensors involved, which usually implies a tedious and burdensome process. We present a method to calibrate the extrinsic parameters of any pair of sensors involving LiDARs, monocular or stereo cameras, of the same or different modalities. The procedure is composed of two stages: first, reference points belonging to a custom calibration target are extracted from the data provided by the sensors to be calibrated, and second, the optimal rigid transformation is found through the registration of both point sets. The proposed approach can handle devices with very different resolutions and poses, as usually found in vehicle setups. In order to assess the performance of the proposed method, a novel evaluation suite built on top of a popular simulation framework is introduced. Experiments on the synthetic environment show that our calibration algorithm significantly outperforms existing methods, whereas real data tests corroborate the results obtained in the evaluation suite. Open-source code is available at \url{https://github.com/beltransen/velo2cam_calibration}.
\end{abstract}


%
\IEEEpeerreviewmaketitle

\section{Introduction}
\label{sec:introduction}
\IEEEPARstart{A}{utonomous} driving relies on accurate information about the environment to make proper decisions concerning the trajectory of the vehicle. High-level inference modules receive these data from the perception systems, which must be therefore endowed with exceptional 
robustness under different circumstances such as illumination and weather.

Consequently, the design of perception systems intended for onboard automotive applications is currently geared towards topologies with several complementary sensory modalities. Vision systems are frequent in close-to-market vehicle setups \cite{Franke2013} due to their ease of integration and their ability to provide appearance information. Stereo-vision systems, which use a pair of cameras separated a fixed distance to get depth information about the environment, stand out as a cost-effective solution able to provide additional dense 3D information to model the surroundings of the vehicle. 

On the other hand, the remarkable development of 3D laser scanning technology has enabled its widespread use in both research and industry driving applications in recent years. Unlike vision systems, LiDAR range measurements are accurate and, frequently, provide information in a full 360\si{\degree} field of view. Setups made of more than one LiDAR device are becoming more and more popular since they allow gathering high-resolution data using compact setups.

\begin{figure}[t]
\centering
    \includegraphics[width=0.95\linewidth]{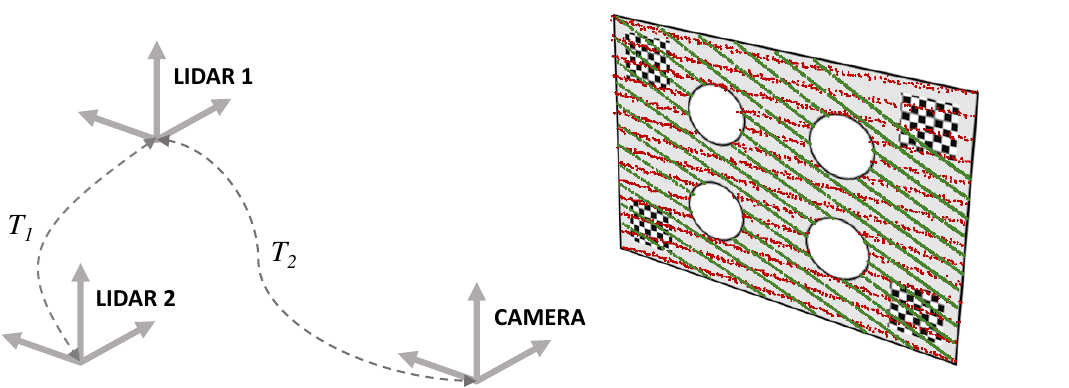}
    \caption{Sample calibration scenario for an arbitrary setup with a camera and two LiDAR scanners, where the calibration target is placed in the overlapping field of view of the involved sensors.}
\label{fig:sidesetup}
\end{figure}

Due to the particular features of these sensory technologies, they are suitable to be part of the same perception system, providing complementary information. In that kind of design, data from the different sensors must be appropriately combined before inference making use of fusion techniques \cite{Feng2019, Beltran2020}. In the most usual setup, sensors have overlapping fields of view (as in Fig.~\ref{fig:sidesetup}), and the advantages conferred by their joint use come from the ability to make correspondences between both data representations. This is the case, for example, with popular multi-modal 3D object detection approaches such as F-PointNet \cite{Qi2017a} or AVOD \cite{Ku2018}. These methods assume that an accurate estimate of the relative pose between the sensors, given by their extrinsic parameters, has been obtained beforehand through a calibration process. 

However, multi-modal calibration is a problem that is still far from trivial. Existing calibration methods suffer from different problems, such as the need for burdensome ad-hoc environments or the lack of applicability to custom sensor setups. It is noteworthy that automotive setups require extraordinary accuracy in the calibration so that it is still valid for data association at long distances.

In this work, we present an original self-calibration method tailored to automotive sensor setups composed of vision devices and multi-layer LiDAR scanners. The approach, preliminarily introduced in \cite{Guindel2017}, comprises robust reference point extraction branches, specific for each modality, and a final 3D point registration stage where the optimal transform relating a pair of sensors is obtained. The proposed solution has various novel properties:
\begin{itemize}
    \item Instead of focusing on a particular setup, our method delivers satisfactory performance with a diversity of cameras and multi-layer LiDAR scanners, even those with a lower resolution (e.g., 16-layer devices). Besides, there are no significant restrictions on the relative pose of the sensors other than the need for an overlap zone between their fields of view: large displacements and strong rotations are admissible.
    \item It also provides a general framework that enables the extrinsic calibration of any combination of camera and LiDAR sensors; not only camera-LiDAR setups but also camera-camera and LiDAR-LiDAR combinations.
    \item A novel fiducial calibration target is introduced to avoid ambiguity and allow an uncomplicated and fast calibration process even in cluttered scenarios. Human intervention is limited to a minimum.
    \item A comprehensive set of experiments shows that the accuracy of the calibration result exceeds other approaches in the literature, being suitable for self-driving applications.
\end{itemize}

Along with this calibration method, we also propose a novel framework to assess extrinsic calibration algorithms based on a simulation environment. This approach provides a perfect ground truth of the transform between sensors in space and establishes a fair benchmark for comparing calibration methods through metrics that truly represent the accuracy of the final estimation. Besides, it allows testing virtually unlimited sensor devices and relative poses to guarantee the generality of the results.

The implementation of the method has been made publicly available to promote reproducibility and provide researchers and practitioners in the field with a convenient tool to face the usual problem of extrinsic calibration.
The software is available as a package in the popular ROS framework\footnote{\url{http://wiki.ros.org/velo2cam_calibration}}. The synthetic test suite used for the experimentation has also been released\footnote{\url{https://github.com/beltransen/velo2cam_gazebo}}.

The remainder of this paper is organized as follows. In Section~\ref{sec:related}, a brief review of related work is provided. Section~\ref{sec:approach} presents a general overview of the proposed algorithm. In Sections~\ref{sec:targetsegmentation} and \ref{sec:registration}, the details of the different stages of the approach are described. Section~\ref{sec:experiments} provides experimental results that assess the performance of the method. Finally, conclusions and open issues are discussed in Section~\ref{sec:conclusion}. 

\section{Related Work}
\label{sec:related}
The issue of calibration of extrinsic parameters expressing the relative pose of sensors of different modalities has been addressed by many researchers in the past, driven by its frequent application in robotics and automotive platforms. The camera-to-range problem has attracted considerable attention, although multi-camera and, more recently, multi-LiDAR systems have also been a subject of interest in the literature.

Calibration is frequently assumed as a process to be performed in a controlled environment before the regular operation of the perception stack. Traditional methods require manual annotation to some extent \cite{Scaramuzza2007}. However, since miscalibrations are common in robotic platforms, research effort has usually focused on automatic approaches. As the process aims to find the correspondence between data acquired from different points of view, unambiguous fiducial instruments have been used as calibration targets, such as triangular boards \cite{Debattisti2013}, polygonal boards \cite{Park2014}, spheres \cite{Pereira2016}, and boxes \cite{Pusztai2017}. Such diversity of shapes deals with the necessity of the targets to be distinguishable in all data representations from sensors. Nonetheless, planar targets are particularly prevalent \cite{Li2011} since they are easily detectable using range information and provide a characteristic shape that can be used to perform geometrical calculations. When monocular cameras are involved, the addition of visual features into the target, such as checkerboards \cite{8917108} or QR markers \cite{2017arXiv170509785D}, allows retrieving the geometry of the scene by inferring the missing scale factor.

With the widespread introduction of LiDAR sensors providing high-resolution 3D point clouds in recent years, research interest has shifted to devices of this kind. Geiger et al. \cite{Geiger2012c} proposed a calibration method based on a single shot in the presence of a setup based on several planar checkerboards used as calibration targets. Velas et al. \cite{Velas2014} proposed an approach enabling the estimation of the extrinsic parameters using a single point of view, based on the detection of circular features on a calibration pattern. A custom calibration target is also used by Zhuang et al. \cite{zhuang2014automatic} to perform calibration between a dense LiDAR scanner and a camera. The method relies on the registration, in the 2D space of the image, of reference points found through elementary processing of both sensors' data. Similarly, Zhou et al. \cite{Zhou2018} made use of a checkerboard to solve the calibration problem by finding correspondences between its representations in LiDAR and image data, using either one or several poses. In general, these methods are targeted to dense range measurements so that 3D LiDAR scanners with lower resolution (e.g., the 16-layer scanner used in this work) entail particular issues that are addressed in this paper. Due to the popularity of this modality, some works are also being devoted to the topic of extrinsic calibration between multiple LiDAR scanners \cite{8814136}.

A relevant second group of approaches dispenses with any artificial calibration targets and uses the features in the environment. Moghadam et al. \cite{Moghadam2013} use linear features extracted from natural scenes to determine the transformation between the coordinate frames. Usually, these methods are suitable for indoor scenes populated with numerous linear landmarks, although some recent works have made efforts to adapt them to outdoor applications \cite{park2020spatiotemporal}. In traffic environments, the ground plane and the obstacles have been used to perform camera-laser calibration \cite{Ponz}, although some parameters are assumed as known. Other approaches are based on semi-automatic methods \cite{Tamas2014} that perform registration on user-selected regions. More recently, Schneider et al. \cite{Schneider2017} took advantage of a deep convolutional neural network to perform all the calibration steps in a continuous online procedure. CalibNet \cite{Iyer2018} has been proposed as a self-supervised calibration framework where the network is trained to minimize the geometric and photometric errors. However, models of this type are difficult to apply to custom sensor setups as they require prior training.

On the other hand, the assessment of calibration methods remains an open issue, given that an accurate ground truth of the parameters defining the relationship between the pose of the sensors cannot be obtained in practice. The lack of standard evaluation metrics has led to the use of custom schemes, which are difficult to extend to other domains and eventually based on inaccurate manual annotations. In this regard, Levinson and Thrun \cite{Levinson2013} presented a method to detect miscalibrations through the variations in an objective function computed from the discontinuities in the scene. A different approach was proposed by Pandey et al. \cite{Pandey2015}, who performed calibration through the maximization of mutual information computed using LiDAR reflectivity measurements and camera intensities.

Nevertheless, current calibration methods still do not provide a comprehensive response to the need to estimate the extrinsic parameters of certain sensor setups, such as the ones found in autonomous driving. They are either excessively focused on specific configurations, lacking generalization ability, or have not been sufficiently validated due to the unavailability of objective assessment methods. We intend to provide a wide-ranging approach able to perform calibration in a large variety of setups and situations, including those usually overlooked, and prove its adequacy quantitatively through a novel benchmark that allows fair comparison with existing methods. 

\section{Method Overview}
\label{sec:approach}
We present a method to estimate the rigid-body transformation that defines the relative pose between a pair of sensors. Each of these sensors can be a LiDAR scanner, a monocular camera, or a stereo camera, in any possible combination. 

The transformation between the pair of sensors can be defined by a vector of six parameters $\boldsymbol{\theta} = (t_x, t_y, t_z, r_x, r_y, r_z)$, which describe the position and rotation of one of the devices in the reference frame attached to the other one. Rotations around the axes ($r_x, r_y, r_z$) are usually referred to as roll, pitch, and yaw angles.

Parameters in $\boldsymbol{\theta}$ unambiguously define a matrix $\mathbf{T}$ that can be used to transform a 3D point between the two coordinate systems. For instance, in a LiDAR-monocular setup, a point $\mathbf{p}_M$ in monocular coordinates, $\left\{M\right\}$, can be transformed into LiDAR space, $\left\{L\right\}$, by means of $\mathbf{p}_L = \mathbf{T}_{LM}\mathbf{p}_M$ once the transformation matrix $\mathbf{T}_{LM}$ is built. Note that, in that particular case, the parameters $\boldsymbol{\theta}_{LM}$, used to obtain $\mathbf{T}_{LM}$, express the pose of $\left\{M\right\}$ with respect to $\left\{L\right\}$.

\begin{figure}[htb]
\centering
    \subfloat[]{%
        \includegraphics[width=0.48\linewidth]{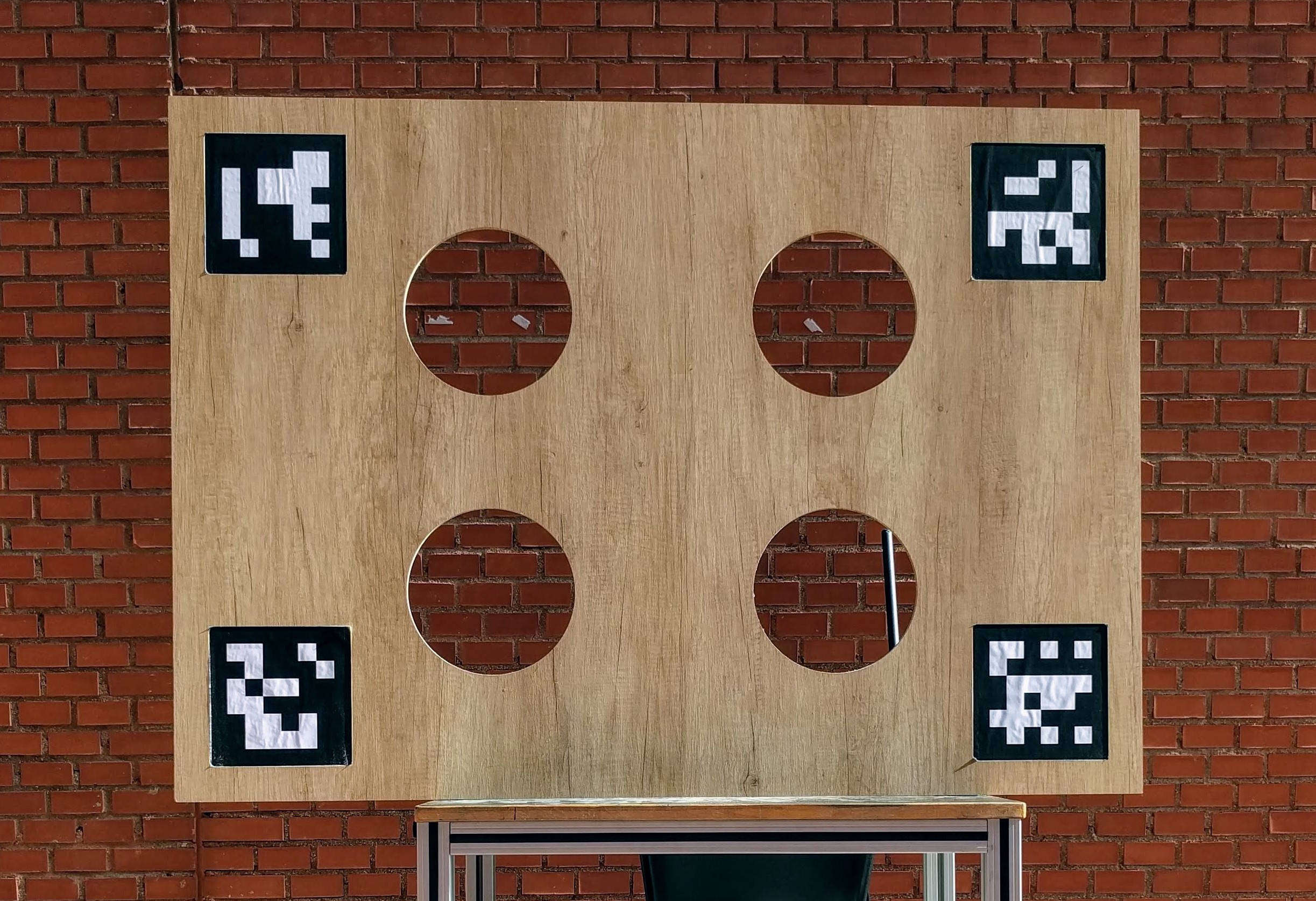}
        \label{fig:target1}
    }
    \subfloat[]{%
        \includegraphics[width=0.48\linewidth]{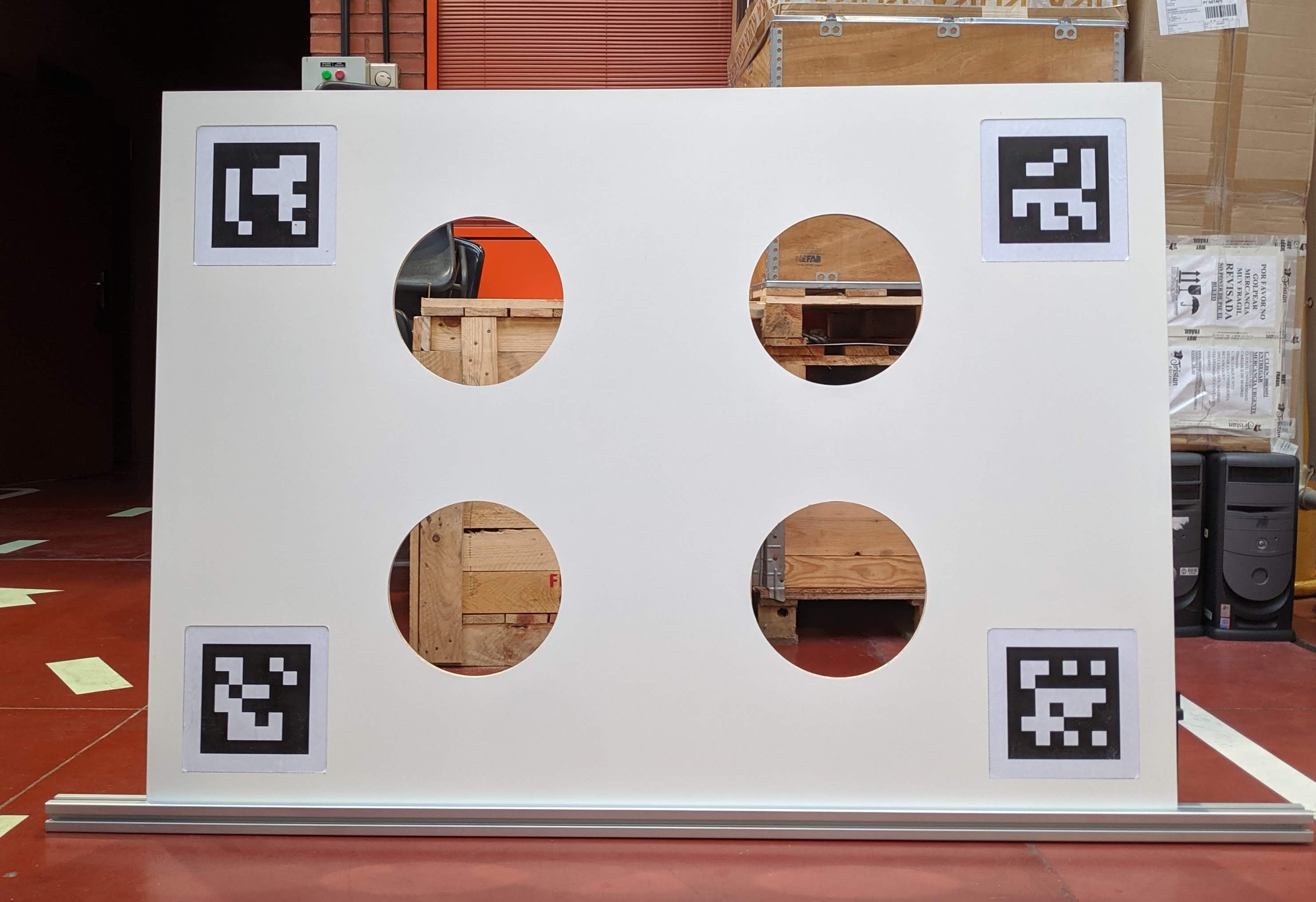}
        \label{fig:target2}
    }
    \caption{Two different embodiments of the custom calibration pattern made with a CNC machine.}
    \label{fig:target}
\end{figure}

With the proposed approach, the transformation is obtained automatically from data retrieved by the sensors to be calibrated. A custom-made planar target is used to provide features that are detected and paired between both data representations. As noticeable in the two different embodiments shown in Fig.~\ref{fig:target}, this calibration pattern is endowed with geometrical and visual characteristics that enable the estimation of keypoints in LiDAR, stereo, and monocular modalities. On the one hand, four circular holes are used to take advantage of geometrical discontinuities in LiDAR and stereo point clouds. On the other hand, four ArUco markers \cite{garrido2014automatic} are placed near the corners so that 3D information can be inferred from monocular images.

The method does not impose severe limits on the relative pose between the devices and is therefore suitable for sensor setups where the magnitudes of the translation and rotation parameters are substantial. Only two reasonable constraints are required. First of all, there has to be an overlapping area between the sensors' field of view, where the calibration target is to be placed. Secondly, the holes in the pattern must be well visible in the data retrieved by the sensors; in particular, whenever range data is involved in the calibration, each circle must be represented by at least three points. In the case of multi-layer LiDAR sensors, this means that at least two scan planes intersect with each of the circles. Moreover, the parameters intrinsic to each device (e.g., focal lengths or stereo baseline) are assumed known.

The procedure is designed to be performed in a static environment. Although the method can provide a quick estimate of the extrinsic parameters with just one pose of the target, it is possible to increase the accuracy and robustness of the results by accumulating several positions, as will be shown later.

\begin{figure*}[htb]
\centering
    \includegraphics[width=\linewidth]{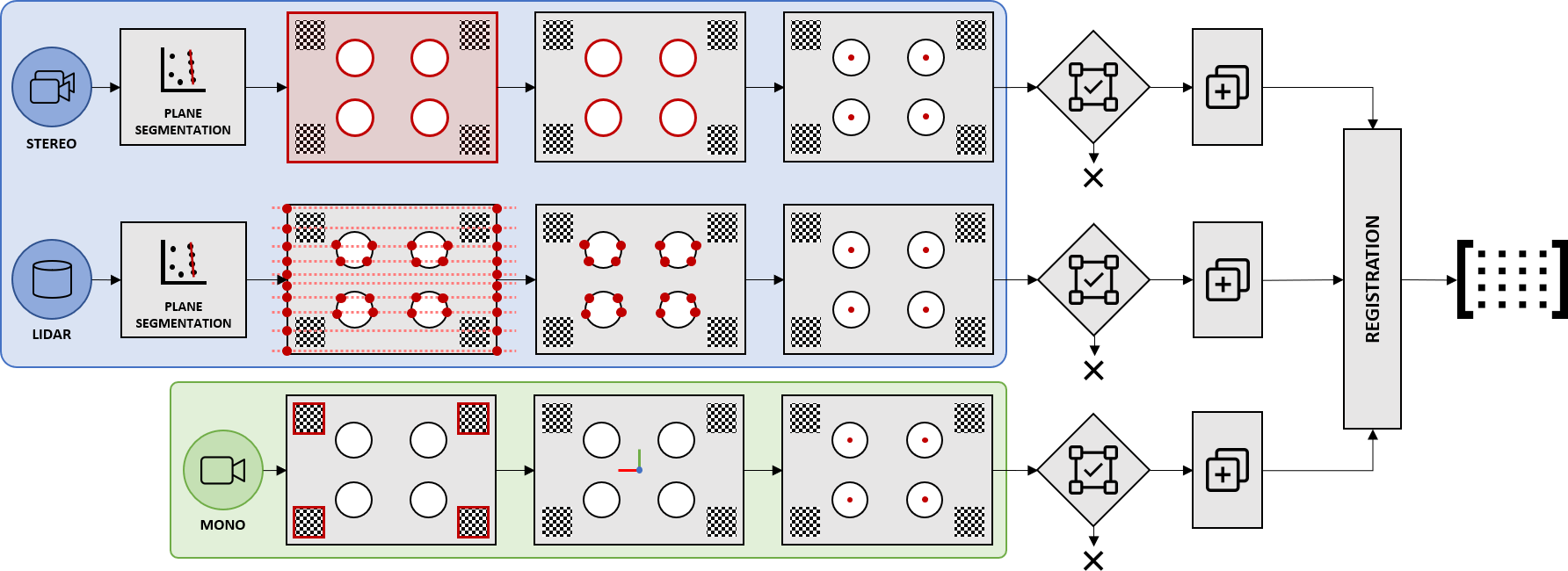}
    \caption{Overview of the different stages of the presented method. For 3D inputs (in blue): plane segmentation, target detection, circles segmentation, and reference points estimation. For monocular cameras (in green): ArUco markers detection, estimation of the target's 3D pose, and reference points estimation. Then, for each frame and modality: geometric consistency check, point aggregation, and sensor registration.}
    \label{fig:approach_schema}
\end{figure*}

The proposed calibration algorithm, illustrated in Fig.~\ref{fig:approach_schema}, is divided into two different stages: the first one involves the segmentation of the calibration target and the localization of the reference points in each of the sensors' coordinate systems; on the other hand, the second one performs the computation of the transformation parameters that enable the registration of the reference points.

\section{Target Segmentation}
\label{sec:targetsegmentation}
This first stage aims to localize the calibration target in each sensor's data. Consequently, the measurements at this stage are relative to the local coordinate system of the corresponding sensor. As the features used to localize the pattern are different for each modality, three different variants of the procedure are proposed here, one per sensor type. In all cases, the output of this stage is a set of four 3D points representing the center of the holes in the target, in local coordinates. These points will be later used to find correspondences between the different data sources.

Although the processing of LiDAR and stereo data has some differences, especially at the beginning of the segmentation stage, both share a common trunk once the useful range data is represented in a 3D point cloud structure. The monocular alternative is substantially different as it relies on the ArUco markers instead. 

The procedure described in this section is intended to be applied to every data frame provided by the corresponding sensor. Data from all sensors are processed in parallel, so they do not have to share a common trigger nor have identical refresh rates, as long as the scene is static.

\subsection{LiDAR Data Preprocessing}
Data from a LiDAR scanner is assumed to be represented as a 3D point cloud, $\mathcal{P}^L_{0}$, with measurements distributed into different layers, as typical in mechanical devices based on rotating mirrors. Before feeding the data to the segmentation procedure, pass-through filters are applied in the three cartesian coordinates to remove points outside the area where the target is to be placed, avoiding spurious detections that could slow down the processing. The limits of the pass-through filters must be set according to the location and size of the sensors' overlapping area. The resulting cloud, $\mathcal{P}^L_{1}$, must represent both the calibration target and the points behind it visible from the LiDAR through the holes. 

As a first step towards segmenting the holes in the pattern, the points representing the edges of the target must be extracted. For the LiDAR modality, we follow the method in \cite{Levinson2013} to find depth discontinuities. Each point in the cloud, $\mathbf{p}_i \in \mathcal{P}_1^L$, is assigned a magnitude representing the depth gradient with respect to their neighbors: 
\begin{equation}
p_{i,\Delta} = \max(p_{i-1, r}-p_{i, r},p_{i+1, r}-p_{i, r},0)
\end{equation}
Where $p_{i, r}$ is the range measurement given by the sensor for the point $\mathbf{p}_i$ (i.e., the spherical radius coordinate), and $\mathbf{p}_{i-1}$ and $\mathbf{p}_{i+1}$ are the points adjacent to $\mathbf{p}_{i}$ in the same scan plane. Then, we filter out all points $\mathbf{p}_i$ with a discontinuity value $p_{i, \Delta} < \delta_{\text{discont},L}$, resulting in $\mathcal{P}^L_{2}$. Note that this procedure assumes that measures from rays passing through the holes exist, so they must collide with some solid located behind the target within the measurement range of the LiDAR.

\subsection{Stereo Data Preprocessing}
When one of the sensors to be calibrated is a stereo-vision system, data processing starts by converting the raw image pair into a 3D point cloud using a stereo matching procedure. In our experiments, we use the Semi-Global Block Matching (SGBM) variant of \cite{Hirschmuller2008} implemented by OpenCV, which we found reasonably accurate for depth estimation. Note that, when this modality is involved, the calibration target is expected to have some texture (e.g., wood grain) so that the stereo correspondence problem can be successfully solved. However, in our experiments, we found that the intensity differences caused by the pattern borders themselves are generally sufficient.  Since the system is assumed canonical and the baseline between cameras known, points can be then provided with an estimate of their depth, and a 3D point cloud $\mathcal{P}^S_{0}$ can be straightforwardly obtained using the pinhole model.   

Similar to the LiDAR branch, pass-through filters are applied to $\mathcal{P}^S_{0}$ to limit the search space. However, for the stereo modality, the extraction of the points representing the target edges in the filtered cloud, $\mathcal{P}^S_{1}$, relies on the appearance information provided by one of the images of the stereo pair. Concretely, a Sobel filter is applied over the image, and then, all points in $\mathcal{P}^S_{1}$ that map to pixels with a low value in the Sobel image (smaller than $\tau_{\text{sobel},S}$) are filtered out, producing $\mathcal{P}^S_{2}$. In this way, edge segmentation is less affected by inaccuracies in border localization, which are frequent in stereo matching.

\subsection{Range Data}
The steps followed to segment the pattern holes in the preprocessed point clouds are common for both the LiDAR and stereo modalities. The intended outcome is an estimate of the 3D location of the centers in sensor coordinates.

\subsubsection{Plane Segmentation}
First of all, a plane segmentation using RANSAC is applied to $\mathcal{P}_{1}$ (the cloud resulting from the pass-through filters, either $\mathcal{P}_{1}^L$ or $\mathcal{P}_{1}^S$), which provides a plane model $\pi$ representing the calibration target. To ensure the model's accuracy, we use a tight RANSAC threshold $\delta_{\text{plane}}$, which neutralizes all the points representing extraneous objects, and impose that the plane must be roughly vertical in sensor coordinates, with a tolerance $\alpha_{\text{plane}}$. If it is impossible to find a plane that fits the data, the current frame is discarded.

Afterward, the plane model $\pi$ is employed in $\mathcal{P}_{2}$ (i.e., the cloud representing the edges of the pattern) to remove all the points not belonging to the plane. A threshold of $\delta_{\text{inliers}}$ is considered for the inliers. Consequently, the new cloud $\mathcal{P}_{3}$ contains only points representing the edges of the calibration target; that is, the outer borders and the holes. 

\subsubsection{Transformation to 2D Space}
As all the remaining points belong to the same plane, dimensionality reduction is performed at this point. This is implemented by transforming $\mathcal{P}_{3}$ so that the XY-plane coincides with $ \pi $ and projecting all the 3D points onto $ \pi $. Points in the resulting $\mathcal{P}_{4}$ cloud are, therefore, in 2D space. 

\subsubsection{Circle Segmentation}
Next, 2D circle segmentation is used to extract a model of the pattern holes present in $\mathcal{P}_{4}$. This step is performed iteratively in a process that seeks out the most supported circle and removes its inliers before starting the search for the next one. Iterations continue until the remaining points are not enough to describe a circle. If at least four circles have been found, the procedure moves forward; otherwise, the current frame is not considered. Inliers are required to be below a threshold of $\delta_{\text{circle}}$ from the model, and only circles within a radius tolerance of $\delta_{\text{radius}}$ are considered.

The points found in the circle segmentation procedure are checked for geometric consistency with the dimensions of the pattern. To that end, the centers are grouped in sets of four, and the dimensions of the rectangle that they form (diagonal, height, width, and perimeter) are compared with the theoretical ones, with a tolerance $\delta_{\text{consistency}}$ expressed as a percentage of deviation from the expected values. Presumably, only one set of centers will fulfill these restrictions; if either none or more than one sets pass the check, the frame is discarded. This step is intended to prune out spurious detections that may occur due to confusion with other elements in the scene.

Once the holes are correctly identified, their centers are converted back from the 2D space defined by $ \pi $ to the 3D space in sensor coordinates, forming the cloud $\mathcal{P}_{p}$. Note that $\mathcal{P}_{p}$ must contain exactly four points. 

\subsection{Monocular Data}
If the sensor to be calibrated is a monocular camera, the extraction of the reference points requires the detection of ArUco markers, which provide the cues necessary to retrieve the geometry of the target.

ArUco markers are synthetic square markers made of a black border and an inner binary matrix designed to allow its unequivocal identification \cite{garrido2014automatic}. In our calibration target, four ArUco markers are used, one on each corner; due to this location, they do not affect either target or hole detection by other modalities.

As both the camera's intrinsic parameters and the marker dimensions are known, it is possible to retrieve the 3D pose of each marker with respect to the camera through the resolution of a classic perspective-n-point (PnP) problem. In our implementation, we handle our four-marker setup as an \textit{ArUco board}, which allows estimating the pose of the calibration target accurately by using all the markers jointly. An iterative Levenberg-Marquardt optimization is carried out to find the board pose that minimizes the reprojection error \cite{Szeliski2010}, using the average pose of the four individual markers as an initial guess. As a result, the 3D position of the center of the board is obtained, along with its orientation in space. 

To generate a set of four points equivalent to the $\mathcal{P}_{p}$ clouds obtained from range data, we extract the points representing the center of the reference holes by taking advantage of the fact that their relative positions in the calibration target are known. These points constitute the resulting cloud $\mathcal{P}_{p}^M$. 

\subsection{Point Aggregation and Clustering}
At the end of the segmentation stage, two clouds $\mathcal{P}_{p}$ must have been generated, one per sensor involved in the calibration. Each represents the 3D location of the reference points (the centers of the target holes) for a single static scene in the coordinate frame attached to the respective sensor.

These data would be enough to find the transform representing the relative pose of the sensors. However, different sources of error inherent to the method (e.g., sensor noise, sparsity of data, and non-deterministic procedures such as RANSAC) can affect the accuracy of the result. To increase the robustness of the algorithm, we augment the information available by repeatedly applying the segmentation step and accumulating the results in two different ways.

\subsubsection{Accumulation over Several Data Frames} 
\label{subsub:clustering}
Since it is usually feasible to maintain the calibration scene static for a certain period, we accumulate the points that compose $\mathcal{P}_{p}$ over $N$ data frames to generate $\mathcal{P}_{p}'$ and then perform Euclidean clustering on this cumulative cloud. If more than four clusters are found, data is considered unreliable and not used for registration; otherwise, cluster centroids, stored in the resulting cloud $\mathcal{P}_{c}$, are employed as a consolidated estimate of the centers' locations. The clustering parameters, namely cluster tolerance $\delta_{\text{cluster}}$, minimum cluster size $N_{\text{cluster}, \text{min}}$, and maximum cluster size $N_{\text{cluster}, \text{max}}$, depend on the number of iterations taken into account. 

According to the experimental results shown later, we usually adopt $N=30$, which offers satisfactory results in a limited timeframe. Naturally, the time necessary to complete the procedure depends on the sensor's framerate but is rarely longer than a few seconds. 

\subsubsection{Accumulation over Several Target Poses}
\label{subsub:multipose}
As will be shown later, the method can deliver an estimated calibration with a single target position. However, it is possible to increase the accuracy of the estimation by considering more than four reference points. If the segmentation procedure is repeated for $M$ different poses of the calibration target with respect to the sensors, the $\mathcal{P}_{c}$ clouds obtained with each pose are accumulated in a $\mathcal{P}_{c}'$ cloud where $4\times M$ reference points are available to perform the registration stage. For the segmentation of each pose, both the sensor and the target are assumed static.

If the poses of the target are selected so that the resulting reference points are not coplanar and cover a wide range of distances from the sensors, the additional constraints provided by the new poses solve possible ambiguities and improve the overall quality of the final calibration.

\section{Registration}
\label{sec:registration}
As a result of the segmentation stage, two clouds $\mathcal{P}_{c}'$, one per sensor, are obtained. They contain the estimated 3D location of the centers of the circles expressed in sensor coordinates; that is, with respect to a frame attached to the sensor. 

The goal of the registration step is to find the optimal parameters $\boldsymbol{\hat{\theta}}$ so that when the resulting transformation $\mathbf{\hat{T}}$ is applied, it results in the best alignment (i.e., minimum distance) between the reference points obtained from both sensors. Note that the approach has been designed to handle only two sources at a time so that the problem can be viewed as a multi-objective optimization with $4\times M$ objective functions. 

Before that, the registration procedure needs that each point in one of the $\mathcal{P}_{c}'$ clouds is correctly paired with its homologous in the other cloud; that is, pairs of points representing the same reference points in both clouds must be associated.

\subsection{Point Association}
A point association procedure has been developed to avoid assuming that reference points in both $\mathcal{P}_{c}$ clouds have the same ordering in their respective coordinate frames. Note that this condition would not be fulfilled when calibrating a front-facing 360\si{\degree} LiDAR and a rear-looking camera, for instance.

Therefore, we convert the four centers in each $\mathcal{P}_{c}$ to spherical coordinates and only assume that the point that appears highest in the cloud, that is, the one with the lowest inclination angle, belongs to the upper row of the calibration target (i.e., either the top-left or the top-right circle).  

Distances from this point to the other three determine the correct ordering. In that way, each point can be associated with the circle in the calibration target that it represents: top-left ($tl$), top-right ($tr$), bottom-left ($bl$), and bottom-right ($br$). The procedure is repeated for each of the $M$ poses of the calibration target, so that each point $\mathbf{p}_i$ in $\mathcal{P}_{c}'$ is provided with labels $p_{i, a}$ and $p_{i, m}$ containing the hole in the pattern and the pose to which it corresponds, respectively:

\begin{alignat}{2}
& p_{i, a} &\in& \left\{tl, tr, bl, br \right\}  \\
& p_{i, m} &\in& \left\{1,\dots, M \right\} 
\end{alignat}

\subsection{Solution}
Later, the two resulting clouds, obtained from two arbitrary modalities $X$ and $Y$ and denoted here by $\mathcal{P}_{c}'^{X}$ and $\mathcal{P}_{c}'^{Y}$, undergo a Umeyama registration procedure \cite{umeyama1991least}, responsible for finding the rigid transformation that minimizes the distance between their corresponding points. That is, assuming that the points in each cloud, $\mathbf{p}_i^{X} \in \mathcal{P}_{c}'^{X}$ and $\mathbf{p}_i^{Y} \in \mathcal{P}_{c}'^{Y}$, are ordered so that, $\forall i$:

\begin{equation}
p_{i, a}^X = p_{i, a}^Y \wedge p_{i, m}^X = p_{i, m}^Y
\end{equation}

Then, the desired transformation matrix $\mathbf{\hat{T}}_{XY}$ is the one that minimizes the least-squares error criterion given by:

\begin{equation}
\frac{1}{4 \cdot M} \sum_{i=1}^{4 \cdot M} \| \mathbf{p}_i^{X} - \mathbf{T}_{XY}\mathbf{p}_i^{Y} \|^2
\end{equation}

This optimization problem is solved through singular value decomposition (SVD) and provides a closed-form solution from which the set of parameters expressing the relative position between both sensors, $\boldsymbol{\hat{\theta}}_{XY}$, can be straightforwardly retrieved. Conveniently, the Umeyama method handles singular situations where all the points are coplanar, as is the case when a single pattern position ($M=1$) is used, thus avoiding misjudging them as reflections.


\section{Experiments}
\label{sec:experiments}
The validation of the proposed approach has been addressed from two different perspectives. First, tests on a realistic synthetic test suite have been performed to retrieve plentiful quantitative data with respect to perfect ground truth. Second, the method has also been applied in a real environment to prove the validity of the approach in real use cases.

All the experiments were carried out without user intervention, except for the tuning of the pass-through filters mentioned in Sec.~\ref{sec:targetsegmentation}, which must be coarsely adapted to the location of the calibration pattern. The rest of the parameters were set to a fixed value for all the experiments, as reported in Table~\ref{tab:parameters}. Unless otherwise stated, reference points are accumulated over 30 frames ($N=30$); however, it should be noted that every frame delivered by the sensors counts toward this limit, regardless of whether a four-point solution has been extracted from it. Conversely, only successful frames ($N'$) are taken into account for the cluster size limits.

\begin{table}[htb] 
	\caption{Setting of Constant Parameters in the Method}
	\label{tab:parameters}
	\centering
	\begin{tabular}{l  l}
		\toprule
    Parameter & Description \\
    \midrule
\multicolumn{2}{l}{\textbf{Preprocessing (edge segmentation)}} \\
$\delta_{\text{discont},L} = \SI{10}{\centi\meter}$ & Distance threshold (LiDAR) \\
$\tau_{\text{sobel},S}=128$ & Sobel intensity threshold (stereo) \\
\midrule
\textbf{Plane segmentation} \\
$\delta_{\text{plane}} = \SI{10}{\centi\meter}$ & Distance threshold \\
$\alpha_{\text{plane}} = \SI{0.55}{\radian}$ & Angular tolerance\\
$\delta_{\text{inliers}} = \SI{10}{\centi\meter}$ & Distance threshold for outlier removal \\
\midrule
\textbf{Circle segmentation} \\
$\delta_{\text{circle},L} = \SI{5}{\centi\meter}$ & Distance threshold (LiDAR) \\
$\delta_{\text{circle},S} = \SI{1}{\centi\meter}$ & Distance threshold (stereo) \\
$\delta_{\text{radius}} = \SI{1}{\centi\meter}$ & Radius tolerance (stereo) \\
$\delta_{\text{consistency}} = \SI{6}{\centi\meter}$ & Geometry consistency tolerance \\
\midrule
\textbf{Clustering} \\
$N_{\text{cluster}, \text{min}} = \frac{1}{2}N'$ & Minimum cluster size \\
$N_{\text{cluster}, \text{max}} = N'$ & Maximum cluster size \\
$\delta_{\text{cluster}} = \SI{5}{\centi\meter}$ & Cluster tolerance \\
		\bottomrule
	\end{tabular}
\end{table}   

\subsection{Synthetic Test Environment}
As stated before, the quantitative assessment of the set of extrinsic parameters relating two sensors in space is a nontrivial issue, as it is impossible, in practice, to obtain exact ground truth. Most works dealing with extrinsic calibration in the literature use manual annotations \cite{Geiger2012c} or other approximations such as scene discontinuities \cite{Levinson2013}.

In order to provide a comprehensive set of data describing the performance of the proposed method, we use the synthetic test suite proposed in \cite{Guindel2017}, where the exact-ground truth of the relative transformation between sensors is available. The open-source Gazebo simulator \cite{Koenig2004} was used, and the operation modes of the three sensor modalities considered in this work (i.e., LiDAR, and stereo and monocular cameras) were faithfully replicated, taking into account the specifications of real devices in terms of field of view, resolution, and accuracy. Table~\ref{tab:sensors} shows the set of devices used in the experiments.
\begin{table}[htb] 
	\caption{Sensor Models used in the Synthetic Environment}
	\label{tab:sensors}
	\centering
	\begin{tabular}{l l l r}
		\toprule
    Device & Modality & Resolution$^{\mathrm{a}}$ & HFOV \\
		 \midrule   
    FLIR Bumblebee XB3 & Stereo & 1280 $\times$ 960 & 43$^{\circ}$\\
    Velodyne VLP-16 & LiDAR & 16 layers, 0.2$^{\circ}$ & 360$^{\circ}$ \\
    Velodyne HDL-32 & LiDAR & 32 layers, 0.2$^{\circ}$  & 360$^{\circ}$ \\
    Velodyne HDL-64 & LiDAR & 64 layers, 0.2$^{\circ}$  & 360$^{\circ}$ \\
    FLIR Blackfly S 31S4C-C & Monocular & 2048 $\times$ 1536 & 85$^{\circ}$ \\
		\bottomrule
	
	\multicolumn{4}{p{230pt}}{$^{\mathrm{a}}$ Image resolution, for cameras, and number of channels and horizontal (azimuth) angular resolution, for LiDAR scanners.}\\
	\end{tabular}
\end{table}       

Remarkably, the different LiDAR devices employed in the experiments are fairly representative of the diversity of laser scanners available in the market regarding the number of scanning layers and their distribution, thus enabling the assessment of the adaptability of the reference point extraction approach.

A model of the fiducial calibration target was also created by mimicking the appearance of the actual wooden embodiment shown in Fig.~\ref{fig:target1}. In the experiments, the target was placed with a wall behind so that LiDAR beams going through the circular holes reach a surface, generating the necessary gradient between foreground and background points. 

Gaussian noise $\epsilon \sim \mathcal{N}(0, (K \sigma_0)^2)$ was applied to the sensors' captured data, with $\sigma_{0}=0.007$ and $\sigma_{0}=\SI{0.008}{\meter}$ for the pixel intensities (expressed in a range from 0 to 1) and the LiDAR distances, respectively. The noise factor $K$ allows simulating ideal, noise-free environments ($K=0$), realistic environments ($K=1$), and noisy environments ($K=2$). $K=1$ is used by default. 

Despite the eventual domain gap, experiments in this controlled setup enable systematic analysis and provide valuable insight into the method that will be otherwise unfeasible. Experimentation in the synthetic suite can be divided into three different focus points: reference point extraction, calibration with a single target position, and multi-pose calibration.

\subsubsection{Single-Sensor Experiments}
The first set of tests is aimed to analyze the accuracy in the extraction of the reference points from the four circular openings in the calibration target. Four different relative positions between sensor and calibration pattern, combining translations and rotations, were considered. Table~\ref{tab:singlesensorsetups} shows the position of the calibration pattern in sensor coordinates for each of these configurations, assuming that axes are defined as customary in LiDAR devices; i.e., $x$ pointing forward, $y$ to the left, and $z$ upward. As in Sec.~\ref{sec:approach}, translation is denoted by $(t_x, t_y, t_z)$, whereas $(r_x, r_y, r_z)$ represent roll, pitch, and yaw rotations (in radians).
\begin{table}[htb] 
	\caption{Relative Sensor-Target Poses for Reference Point Extraction Assessment}
	\label{tab:singlesensorsetups}
	\centering
	\begin{tabular}{l r r r r r r r } 
		\toprule
		& \multicolumn{4}{c}{Translation (m)} & \multicolumn{3}{c}{Rotation (rad)} \\
		\cmidrule(lr){2-5} \cmidrule(lr){6-8}
    Cfg. & \multicolumn{1}{c}{$t_x$} & \multicolumn{1}{c}{$t_y$} & \multicolumn{1}{c}{$t_z$} & \multicolumn{1}{c}{$\left| \boldsymbol{t} \right|$} & \multicolumn{1}{c}{$r_x$} & \multicolumn{1}{c}{$r_y$} & \multicolumn{1}{c}{$r_z$} \\
		 \midrule   
    P1 & 2.00 & 0.00 & $-$0.50 & 2.06 & 0.0 & 0.0 & 0.0 \\ 
    P2 & 3.63 & $-$0.50 & $-$0.28 & 3.67 & 0.8 & 0.0 & 0.0 \\
    P3 & 5.38 & $-$0.10 & $-$0.50 & 5.41 & 0.0 & $-$0.2 & 0.0 \\
    P4 & 6.50 & $-$1.39 & $-$1.43 & 6.80 & 0.0 & 0.0 & $-$0.4 \\
		\bottomrule
	\end{tabular}
\end{table}      

These setups were purposely chosen to investigate the limits of the reference point extraction branches. In fact, the method was unable to provide results in some extreme configurations; concretely, with the VLP-16 LiDAR in P3 and P4, the HDL-32 LiDAR in P4, and the stereo camera in P4 as well. In the case of the LiDAR scanners, their limited resolution made it impossible to find the circles at far distances, whereas the stereo was affected by the substantial degradation in depth estimation that this modality suffers as the distance increases. In typical use cases, it should be possible to avoid these situations by restricting the pattern locations to a reasonable range of distances with respect to the sensors.

The reference point localization performance was measured by determining the distance between the estimation provided by the approach and the ground-truth position of the center of the corresponding circle. The assignment was unambiguous in all cases and could be straightforwardly performed based on distance. Results were aggregated over three iterations for each pose and modality to account for the effect of the stochastic processes in the pipeline (e.g., RANSAC segmentations).

\begin{figure*}[htb]
\centering
    \subfloat[Stereo]{%
        \includegraphics[width=0.25\linewidth]{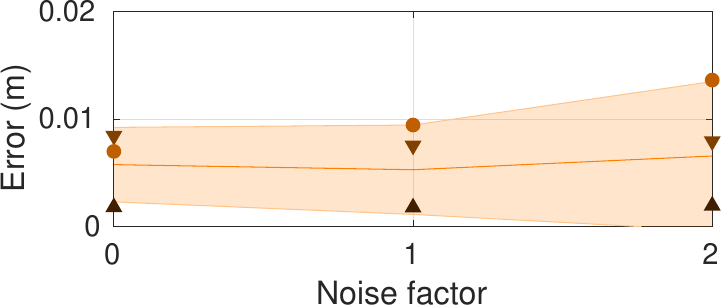}
        \label{fig:ss1}
    }
    \subfloat[VLP-16]{%
        \includegraphics[width=0.25\linewidth]{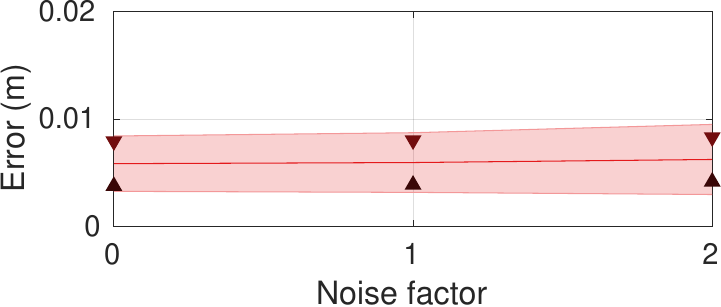}
        \label{fig:ss2}
    }
    \subfloat[HDL-32]{%
        \includegraphics[width=0.25\linewidth]{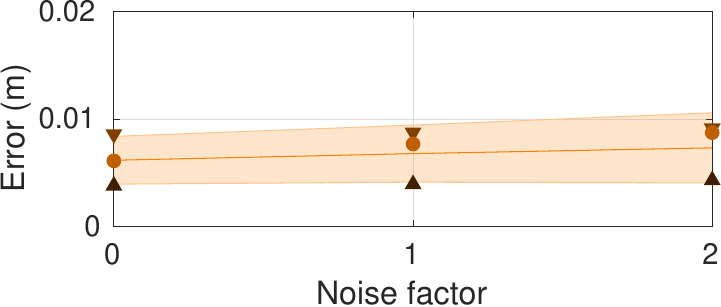}
        \label{fig:ss3}
    }
    \hfill
    \\
    \subfloat[HDL-64]{%
        \includegraphics[width=0.25\linewidth]{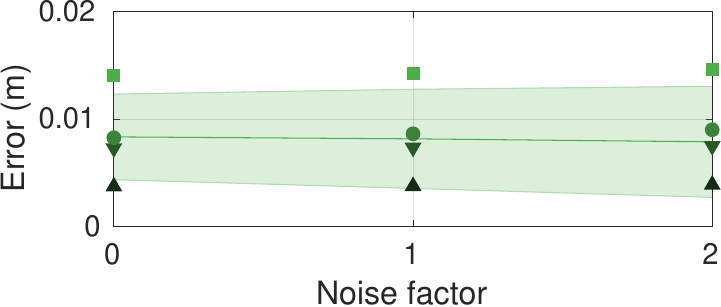}
        \label{fig:ss4}
    }
    \subfloat[Monocular]{%
        \includegraphics[width=0.25\linewidth]{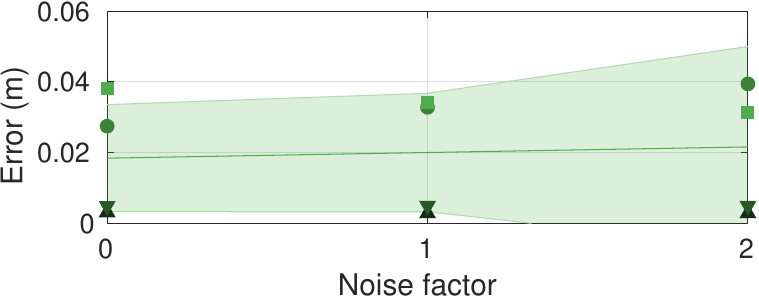}
        \label{fig:ss5}
    }
\caption{Euclidean error in single-frame reference point localization vs. noise level ($K$), for each tested modality. The mean is depicted as a solid line, whereas the shaded area represents the standard deviation. Mean errors for each pose are depicted as individual markers (P1: upwards triangle, P2: downwards triangle, P3: circle, P4: square).}
\label{fig:singlesensor}
\end{figure*}

Firstly, Fig.~\ref{fig:singlesensor} analyzes the effect of noise in the reference points location error. The results show that the procedure is highly robust to noise in all the modalities, given that the impact is limited to an increase in the standard deviation of the error in noisy situations ($K=2$). In all cases, the error is well below \SI{1}{\centi\meter} for the P1 and P2 configurations (upwards and downwards triangle markers in the graph), whereas P3 (circle markers) and, especially, P4 (square markers) involve a significant increase across all the noise levels. This fact is particularly noticeable for the monocular modality (please note the different scale in the y-axis), where the accuracy in the detection of the ArUco markers proves to be much more sensitive to the size of their projections onto the image than to the pixel-wise noise.  

\begin{figure}[htb]
\centering
        \includegraphics[width=\linewidth]{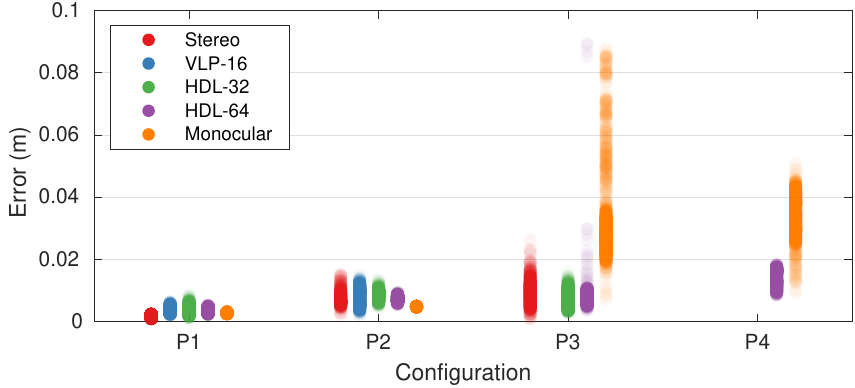}
\caption{Euclidean error in single-frame reference point localization, for each tested configuration and modality, with realistic noise.} 
\label{fig:dispersion}
\end{figure}

Focusing on the realistic noise setup ($K=1$), Fig.~\ref{fig:dispersion} shows the single-frame estimation error in each of the four configurations, further highlighting the relative position between sensor and calibration pattern as a significant factor. Apart from the most challenging configurations, the reference point localization proves accurate and precise across all the modalities, with LiDAR scanners exhibiting high robustness even in P3 and P4. As mentioned before, monocular struggles with these configurations but shows an excellent performance in P1 and P2.

The effect of the point aggregation and clustering strategy introduced in Sec.~\ref{subsub:clustering} is investigated in Table~\ref{tab:rmsesinglecluster}, where the root-mean-square error (RMSE) of single-frame estimations and 30-iterations cluster centroids are compared under realistic noise conditions. The cluster centroid proves to be a consistently better representation of the reference points than the single-frame estimation in all cases, achieving a more remarkable improvement in situations with high dispersion; e.g., stereo in P3 ($25.22\%$ error reduction) or HDL-64 also in P3 ($19.02\%$ error reduction). 

\addtolength{\tabcolsep}{-1.3pt} 
\begin{table}[htb] 
	\caption{RMSE (mm) in Reference Point Location using a Single-Shot Estimation (S) and the Cluster Centroid at $N=30$ (C)}
	\label{tab:rmsesinglecluster}
	\centering
	
	\begin{tabular}{l r r r r r r r r }
		\toprule
		
    & \multicolumn{2}{c}{P1} & \multicolumn{2}{c}{P2} & \multicolumn{2}{c}{P3} & \multicolumn{2}{c}{P4} \\
    \cmidrule(lr){2-3} \cmidrule(lr){4-5} \cmidrule(lr){6-7}  \cmidrule(lr){8-9}
     & \multicolumn{1}{c}{S} & \multicolumn{1}{c}{C} & \multicolumn{1}{c}{S} & \multicolumn{1}{c}{C} & \multicolumn{1}{c}{S} & \multicolumn{1}{c}{C} & \multicolumn{1}{c}{S} & \multicolumn{1}{c}{C} \\
		 \midrule   
    Stereo & 1.84 & 1.83 & 7.82 & 6.83 & 10.11 & 7.56 & {-} & {-} \\ 
    VLP-16 & 3.98 & 3.87 & 8.39 & 8.27 & {-} & {-} & {-} & {-} \\
    HDL-32 & 4.12 & 3.98 & 8.82 & 8.61 & 8.02 & 7.41 & {-} & {-} \\
    HDL-64 & 3.81 & 3.74 & 7.38 & 7.29 & 9.99 & 8.09 & 14.43 & 14.28 \\
    Mono & 2.82 & 2.80 & 4.92 & 4.91 & 35.78 & 35.58 & 34.70 & 33.87 \\
		\bottomrule
	\end{tabular}
\end{table}       
\addtolength{\tabcolsep}{1.3pt}

Once again, the results suggest that the accuracy in reference point extraction is primarily impacted by the relative pose of the calibration target and, to a lesser extent, by the sensor modality. In contrast, the density of LiDAR data seems to have little influence on the results, although minor differences in the way laser beams interact with the target depending on the layer distribution produce a few counterintuitive results.

\subsubsection{Single-Pose Experiments}
Next, the full calibration pipeline will be evaluated considering only a single target position; that is, for $M=1$. To that end, four combinations representative of real automotive sensor setups were analyzed:
\begin{enumerate}[label=\Alph*.]
\item HDL-32/HDL-64 (LiDAR/LiDAR)
\item Monocular/HDL-64  (camera/LiDAR)
\item Monocular/monocular (camera/camera)
\item Stereo/HDL-32 (camera/LiDAR)
\end{enumerate}
Setups A and C embody situations where several devices of the same modality are included in the same sensor setup to enhance the field of view or the resolution of the captured data, whereas setups B and D exemplify setups aimed at camera/LiDAR sensor fusion. Both situations are frequently found in the onboard perception literature, even jointly on the same platform, e.g., \cite{Caesar2019}.

\begin{figure*}[htb]
\centering
    \subfloat[]{%
        \includegraphics[width=0.32\linewidth]{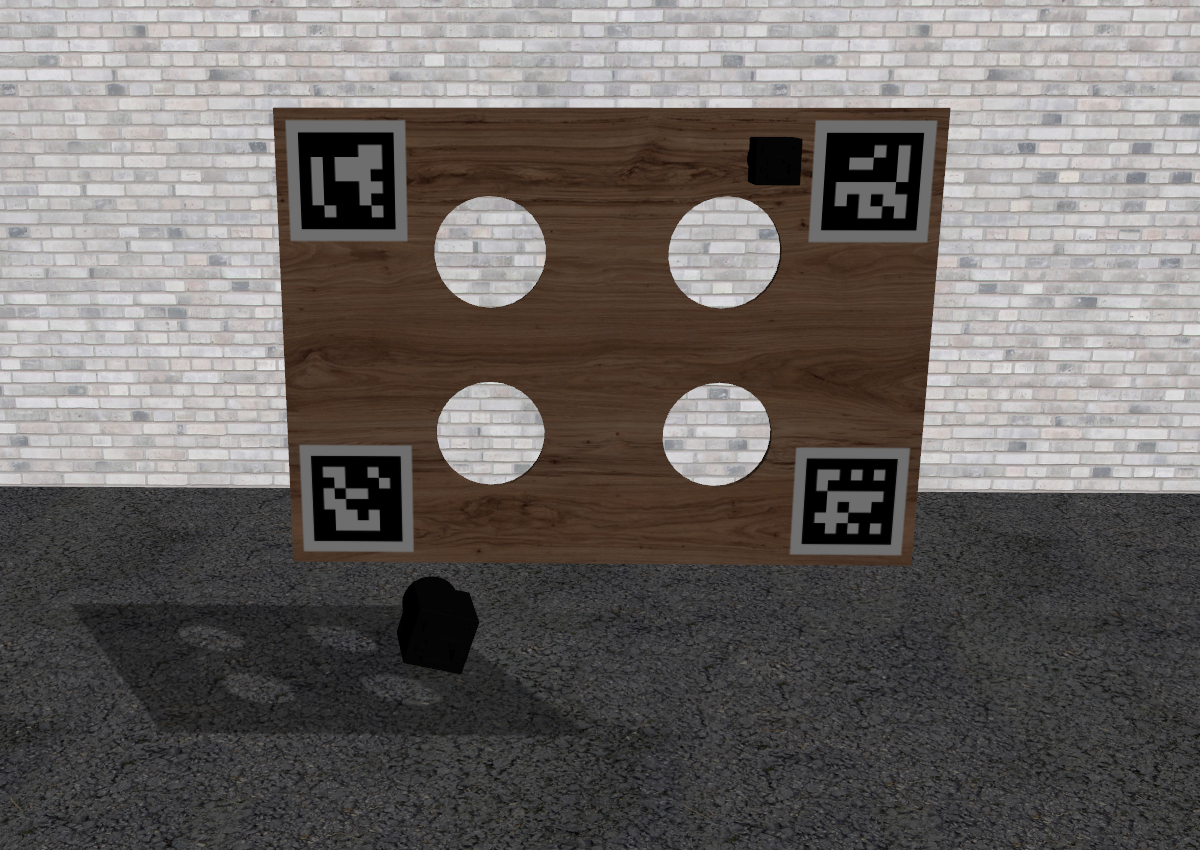}
        \label{fig:pic1}
    }
    \subfloat[]{%
        \includegraphics[width=0.32\linewidth]{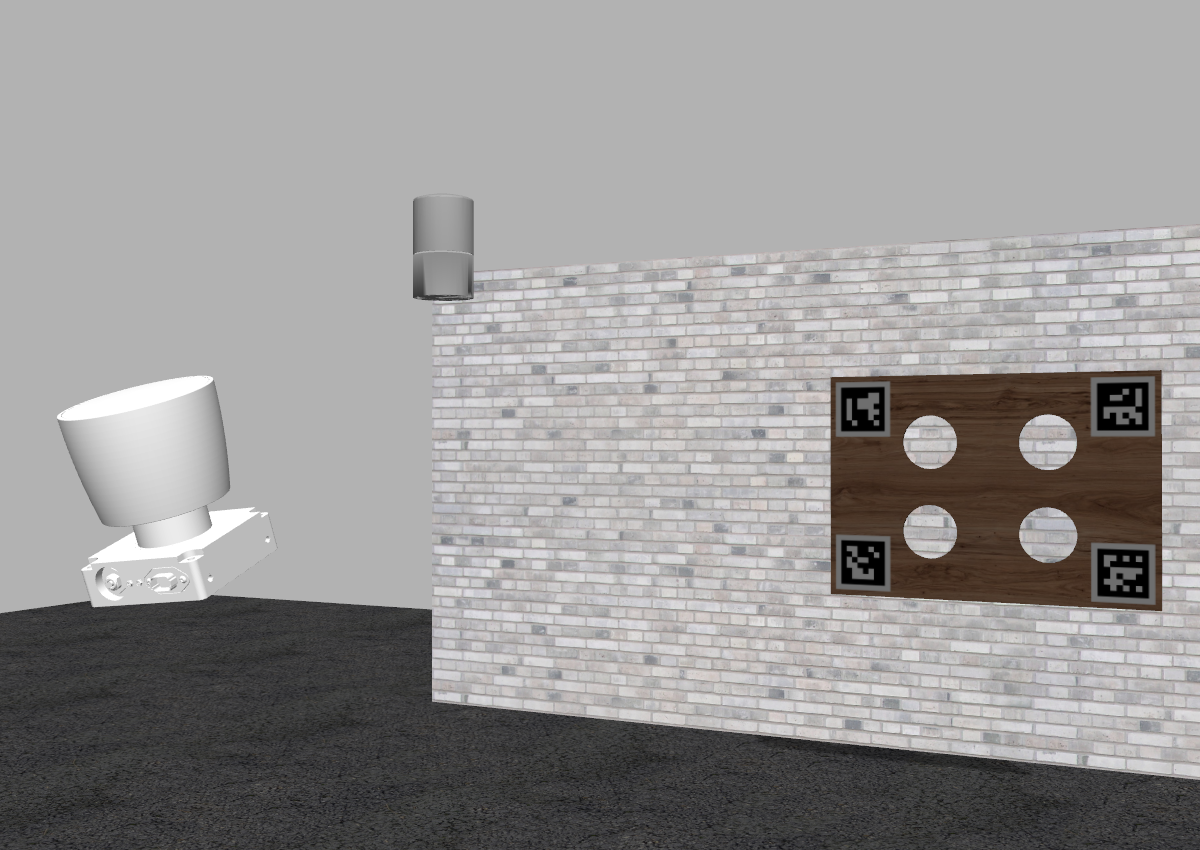}
        \label{fig:pic2}
    }
    \subfloat[]{%
        \includegraphics[width=0.32\linewidth]{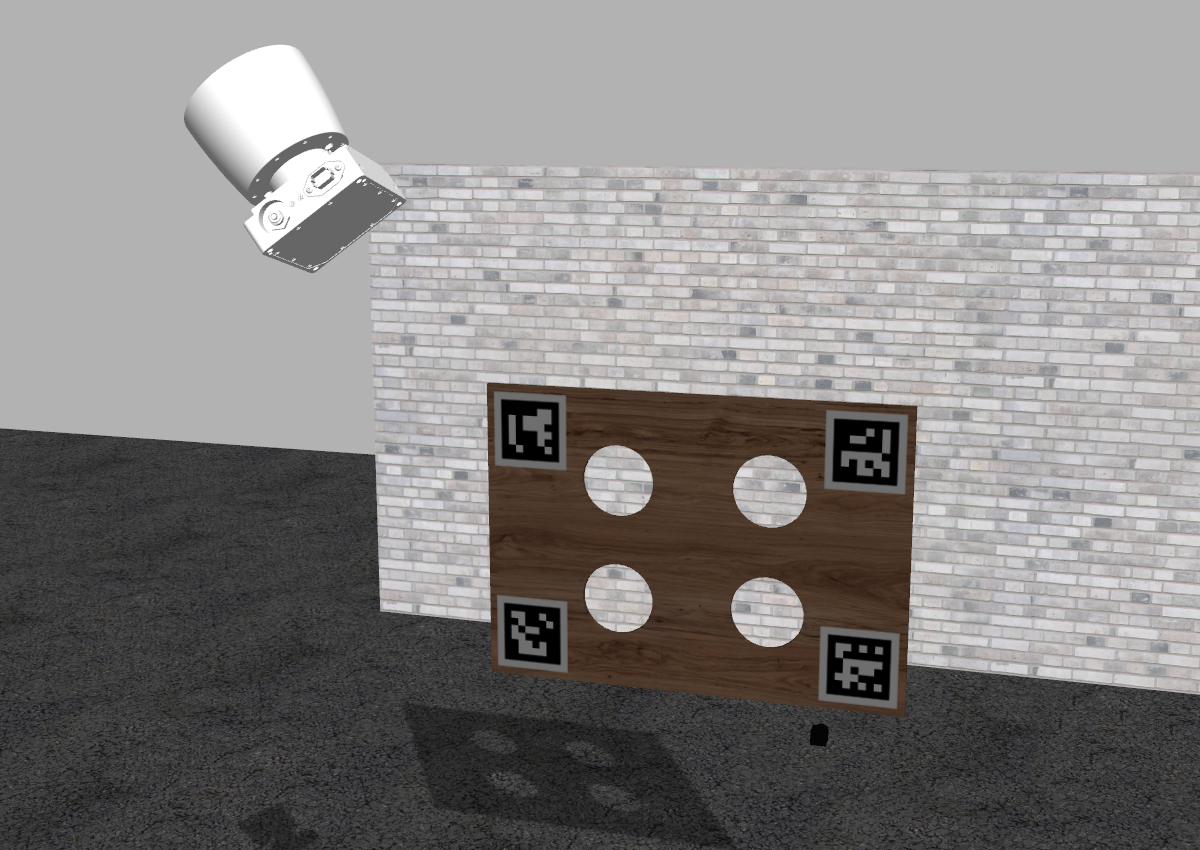}
        \label{fig:pic3}
    }
    \hfill
\caption{Sensor setups for the single-pose experiments in the synthetic environment: P1 (a), P2 (b), and P3 (c)}
\label{fig:picssinglepose}
\end{figure*}

For each setup, the three different relative positions between sensors reported in Table~\ref{tab:setups} were considered. They were picked from \cite{Guindel2017} as a representative set of configurations involving a wide range of translations and rotations. Representative pictures of these configurations in the synthetic test suite are depicted in Fig.~\ref{fig:picssinglepose}. As in the previous case, three different iterations were considered in the results for each possibility. In all cases, the calibration pattern was placed arbitrarily in a location suitable for both sensors. Like in the per-sensor analysis, different distances to the target are used to further study its effect on final calibration.

\begin{table}[htb] 
	\caption{Transformation parameters of the different calibration scenarios}
	\label{tab:setups}
	\centering
	\begin{tabular}{l  r r r r r r   }
		\toprule
    Cfg. & {$t_x$  (m)} & {$t_y$ (m)} & {$t_z$ (m)} & {$\psi$ (rad)} & {$\theta$ (rad)} & {$\phi$ (rad)} \\
		 \midrule   
    P1 & $-$0.300 & 0.200 & $-$0.200 &  0.300 & $-$0.100 & 0.200 \\ 
    P2 & $-$0.128 & 0.418 & $-$0.314 & $-$0.103 & $-$0.299 & 0.110 \\
    P3 & $-$0.433 & 0.845 &  1.108 & $-$0.672 &  0.258 & 0.075 \\
		\bottomrule
	\end{tabular}
\end{table}       

The analysis is now focused on the final calibration result. Therefore, following \cite{Geiger2012c}, results are given in terms of the linear ($e_t$) and angular ($e_r$) errors between the estimated rigid-body transformation and the ground truth:
\begin{align}
e_t &= \| \boldsymbol{\hat{t}}-\boldsymbol{t} \| \label{eq:linear_dif2}\\
e_r &= \angle(\mathbf{\hat{R}}^{-1}\mathbf{R}) \label{eq:rot_dif2}
\end{align}
Where $\boldsymbol{t}$ is the translation vector, $\boldsymbol{t} = (t_x, t_y, t_z)$, and $\mathbf{R}$ the $3 \times 3$ rotation matrix, representing the $r_x$, $r_y$, and $r_z$ rotations; both elements compose the transformation matrix:
\begin{equation}
\mathbf{T} = \begin{bmatrix}
\mathbf{R} & \boldsymbol{t} \\
\boldsymbol{0} & 1
\end{bmatrix} 
\end{equation}

\begin{figure}[htb]
\centering
    \subfloat[]{%
        \includegraphics[width=0.48\linewidth]{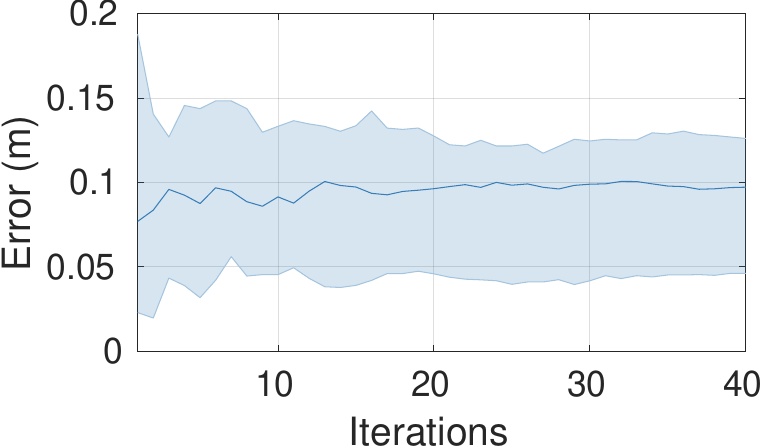}
        \label{fig:test1}
    }
     \subfloat[]{%
        \includegraphics[width=0.48\linewidth]{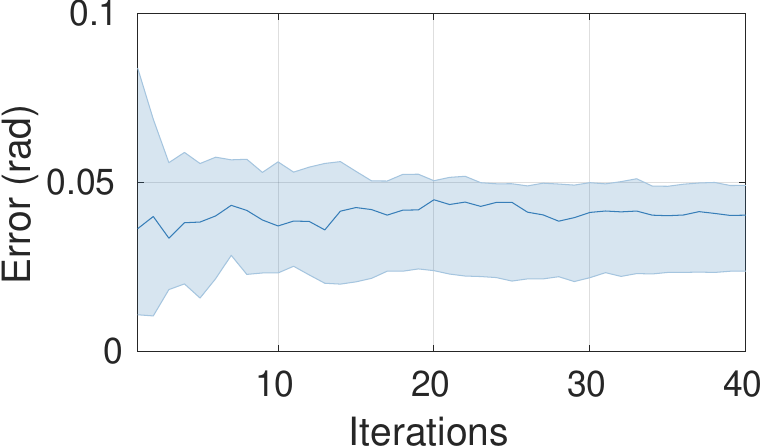}
        \label{fig:test2}
    }
\caption{Linear (a) and angular (b) calibration errors vs. number of iterations considered for clustering ($N$). The solid line represents the median and the shaded area, the interquartile range (IQR).}
\label{fig:singlepose}
\end{figure}

In the first place, the effect of the number of data frames used for reference point extraction, $N$, was studied. Fig.~\ref{fig:singlepose} aggregates the error for every setup and configuration when the calibration procedure is stopped at a point in the $N=\left[1, 40\right]$ interval. The results suggest that the method can provide a reliable estimation of the extrinsic parameters in a wide range of values of $N$, even with very few iterations. Nevertheless, $N=30$ offers a fair accuracy-time tradeoff where outliers are extremely rare.

Table~\ref{tab:singleposeresults} shows the linear ($e_t$) and angular ($e_r$) calibration errors sorted by sensor setup and configuration for $N=30$. 
Monocular/monocular calibration (setup C) shows excellent accuracy and precision, in line with the reference point extraction results, featuring errors up to 100 times smaller than the rest of the setups. On the contrary, the stereo/HDL-32 (setup D) presents higher errors, likely due to the difficulties found by the stereo matching procedure to provide an accurate depth estimation at the distance where the pattern was placed in the experiments. 
Despite this, we observed that the implementation of the ArUco detector in use (OpenCV \textit{aruco} module) was considerably more sensitive to light conditions than the stereo matching approach, so the method based on the stereo modality might still be useful in some instances. 
Overall, the results are reasonably accurate, even though the single-target situation poses a very challenging case for registration due to the coplanarity of the reference points, which can eventually become a source of ambiguity. 

\begin{table}[htb] 
	\caption{Mean (and Std Dev) of Linear ($e_t$) and Angular ($e_r$) Calibration Errors for Different Setups using a Single Target Pose ($M=1$)}
	\label{tab:singleposeresults}
	\centering
	\begin{tabular}{l l r r r} 
		\toprule
		{Set.} & {Error} & \multicolumn{1}{c}{P1} & \multicolumn{1}{c}{P2} & \multicolumn{1}{c}{P3}\\
		 \midrule   
    \multirow{2}{*}{A} & $e_t$ (cm) & 8.94 (1.49) & 17.39 (2.13) & 11.95 (1.56) \\ 
    & $e_r$ ($10^{-2}$ rad) & 4.36 (0.72) & 3.91 (0.48) & 5.80 (0.78) \\
    \midrule
        \multirow{2}{*}{B} & $e_t$ (cm) & 10.34 (0.53) &  4.31 (0.29) & 9.68 (0.22) \\ 
    & $e_r$ ($10^{-2}$ rad) & 5.08 (0.26) & 2.23 (0.13) & 4.74 (0.12) \\
        \midrule
        \multirow{2}{*}{C} & $e_t$ (cm) & 0.17 (0.01) &  0.08 (0.00) & 0.16 (0.00) \\ 
    & $e_r$ ($10^{-2}$ rad) & 0.03 (0.01) & 0.04 (0.00) & 0.04 (0.00) \\
        \midrule
        \multirow{2}{*}{D} & $e_t$ (cm) & 9.62 (1.12) & 47.02 (1.49) & 31.60 (2.95) \\ 
    & $e_r$ ($10^{-2}$ rad) & 2.85 (0.34) & 14.87 (0.47) & 8.75 (0.84) \\
		\bottomrule
	\end{tabular}
\end{table}   
  
Table~\ref{tab:singleposecomparison} shows a comparison of the proposed approach with two single-pose LiDAR-camera calibration methods in the literature: the one by Geiger et al. \cite{Geiger2012c}, which estimates both the intrinsic and extrinsic parameters of the sensors with only one shot, and the one proposed by Velas et al. \cite{Velas2014}, which makes use of a calibration pattern very similar to ours. For a fair comparison, all the methods were fed with sensor data from the synthetic test suite, as reported in \cite{Guindel2017}. The sensor setup was composed of the stereo camera and the HDL-64 LiDAR introduced in Table~\ref{tab:sensors}. We consider the two available options for reference point extraction in visual data: stereo and monocular, the latter employing the left image of the stereo rig as input. The errors were averaged over the same three poses used in the previous experiments.

\begin{table}[htb] 
	\caption{Mean (and Standard Deviation) of Linear ($e_t$) and Angular ($e_r$) Calibration Errors using a Single Target Pose ($M=1$)}
	\label{tab:singleposecomparison}
	\centering
	\begin{tabular}{l c c} 
		\toprule
     {Method} &  {$e_t$ (m)} & {$e_r$ (rad)} \\
		 \midrule   
		Geiger et al. \cite{Geiger2012c} & 0.93 (0.36) & 1.30 (1.35) \\ 
		Velas et al. \cite{Velas2014}   & 0.99 (1.17) & 0.35 (0.37) \\
		Ours (Stereo-LiDAR)             & 0.12 (0.09) & 0.04 (0.03) \\ 
        Ours (Monocular-LiDAR)          & 0.12 (0.12) & 0.04 (0.03) \\ 
		\bottomrule
	\end{tabular}
\end{table}    

According to these results, the stereo and mono alternatives yield similar accuracy, significantly outperforming the other methods. Particularly noteworthy is the substantial improvement in angular error brought about by our approach, which stands out as the only one suitable for data fusion at far distances. These results prove that the baseline method, requiring a single pose of the calibration pattern ($M=1$), works acceptably and provides a solid foundation for the full version with $M>1$.

\subsubsection{Multi-Pose Experiments}
The last set of experiments focuses on the aggregation strategy presented in Sec. \ref{subsub:multipose}, where the registration procedure is performed on $M \times 4$ points coming from $M$ different calibration target positions. The sensor setups are identical to those used in the single-pose tests, but only the first configuration (P1) has been selected. For every sensor pair, the calibration pattern was moved along five different poses within a range of $5 \times 5$ m in front of the devices, up to \SI{6}{\meter} in depth. To avoid the eventual bias introduced by the poses ordering, results are obtained through three different iterations in which the sorting is changed.

\begin{figure}[htb]
\centering
    \subfloat[]{%
        \includegraphics[width=0.48\linewidth]{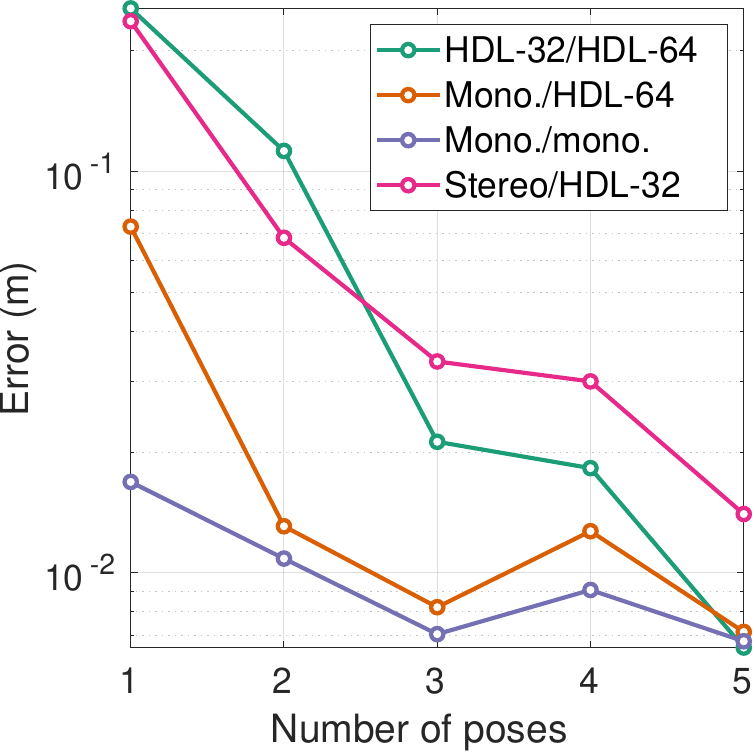}
        \label{fig:multipose1}
    }
    \subfloat[]{%
        \includegraphics[width=0.48\linewidth]{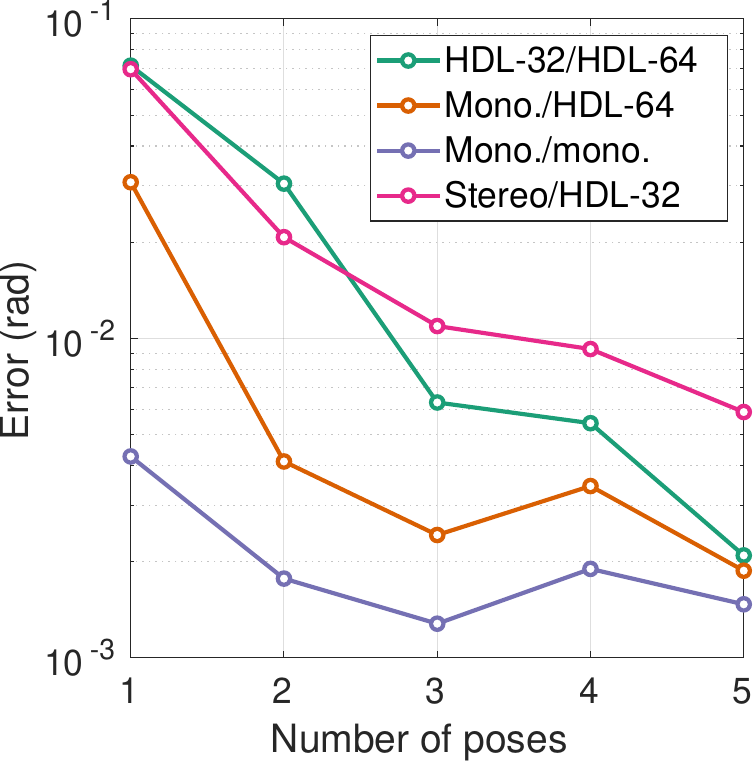}
        \label{fig:multipose2}
    }
\caption{RMSE of the linear (a) and angular (b) calibration errors (m and rad) vs. number of calibration poses ($M$) for four sensor setups.}
\label{fig:multipose}
\end{figure}

The evolution of the linear and angular calibration errors with $M$ follows an almost-exponential decay for all the tested setups, as shown in Fig.~\ref{fig:multipose} (please note the logarithmic scale). Only by introducing an additional target pose, an average reduction of $61.2\%$ (linear) / $68.15\%$ (angular) can be achieved. Increasing the number of poses is positively beneficial up to $M=3$; higher values lead to mixed effects ranging from almost neutral to slightly positive. Nevertheless, when five poses are employed, the average errors drop by $85.42\%$ (linear) / $87.01\%$ (angular). The largest decreases correspond to the HDL-32/HDL-64 setup, where the reduction is around $97\%$ for both kinds of errors, yielding a final calibration with a deviation of \SI{6.5}{\milli\meter} and \SI{0.002}{\radian} from the ground truth.  

The proposed approach has been compared with the state-of-the-art method recently introduced by Zhou et al. \cite{Zhou2018}, aimed at LiDAR-camera calibration using one or several views of a checkerboard. To that end, we used the implementation included in the MATLAB Lidar Toolbox \cite{Matlab}. Tests were performed with the monocular/HDL-64 sensor setup, using $M=2$ and $M=3$ poses of the respective calibration patterns. Mean calibration errors by both methods are shown in Table~\ref{tab:matlabcomparison}.  

\begin{table}[htb] 
	\caption{Mean of Linear ($e_t$) and Angular ($e_r$) Calibration Errors using Several Target Poses ($M>1$)}
	\label{tab:matlabcomparison}
	\centering
	\begin{tabular}{l c c c c }
		\toprule
		 & \multicolumn{2}{c}{$M=2$} & \multicolumn{2}{c}{$M=3$} \\
    \cmidrule(lr){2-3} \cmidrule(lr){4-5} 
     {Method} &  {$e_t$ (cm)} & {$e_r$ ($10^{-2}$ rad)} & {$e_t$ (cm)} & {$e_r$ ($10^{-2}$ rad)} \\
		 \midrule   
        Zhou et al. \cite{Zhou2018} & 1.51 & 0.63 & 1.08 & 0.50 \\ 
        Ours        & 1.15 & 0.39 & 0.82 & 0.24 \\ 
		\bottomrule
	\end{tabular}
\end{table}       

As apparent from the results, the performance of both approaches is comparable, although our method achieves consistent improvements that even exceed 50\% for the angular error when $M=3$. These results confirm the effectiveness of the aggregation of reference points across different target locations, providing a calibration solution that features subcentimeter accuracy.

\subsection{Real Test Environment}
The set of experiments presented in the previous section offers a systematic and exact analysis of the performance of the proposed calibration method. Nevertheless, experiments in a real use case were also carried out to validate the applicability of the approach, assessing its adequacy to meet the requirements of the intended application. 

The CNC manufactured calibration targets shown in Fig.~\ref{fig:target} were employed in the process.
We performed two rounds of experiments using different sensor stacks to test the multiple capabilities of the approach adequately. Both configurations were mounted on an experimental vehicle's roof rack.

\begin{figure}[htb]
\centering
    \subfloat[]{%
        \includegraphics[width=0.48\linewidth]{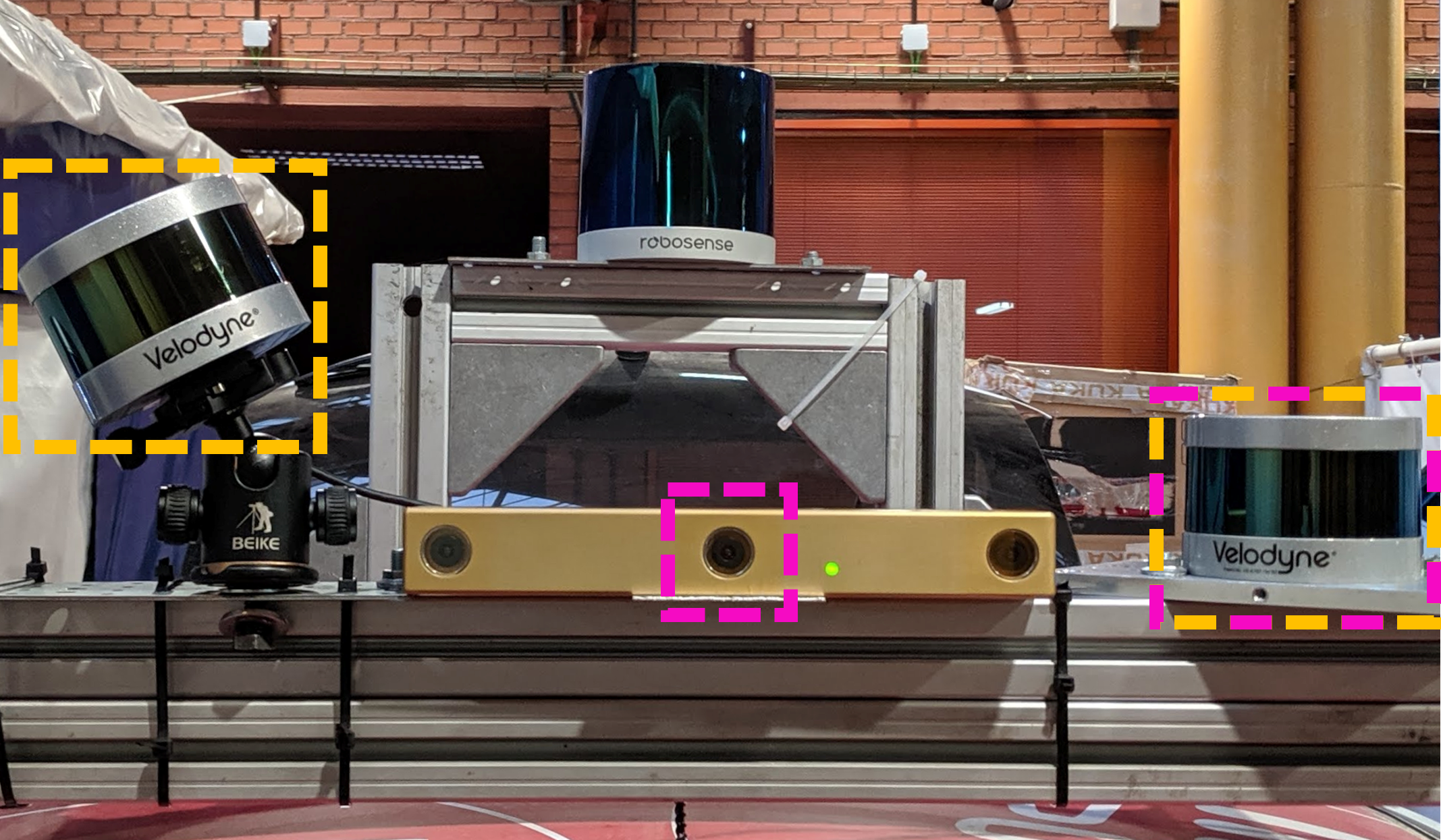}
        \label{fig:realsensors1a}
    }
    \subfloat[]{%
        \includegraphics[width=0.48\linewidth]{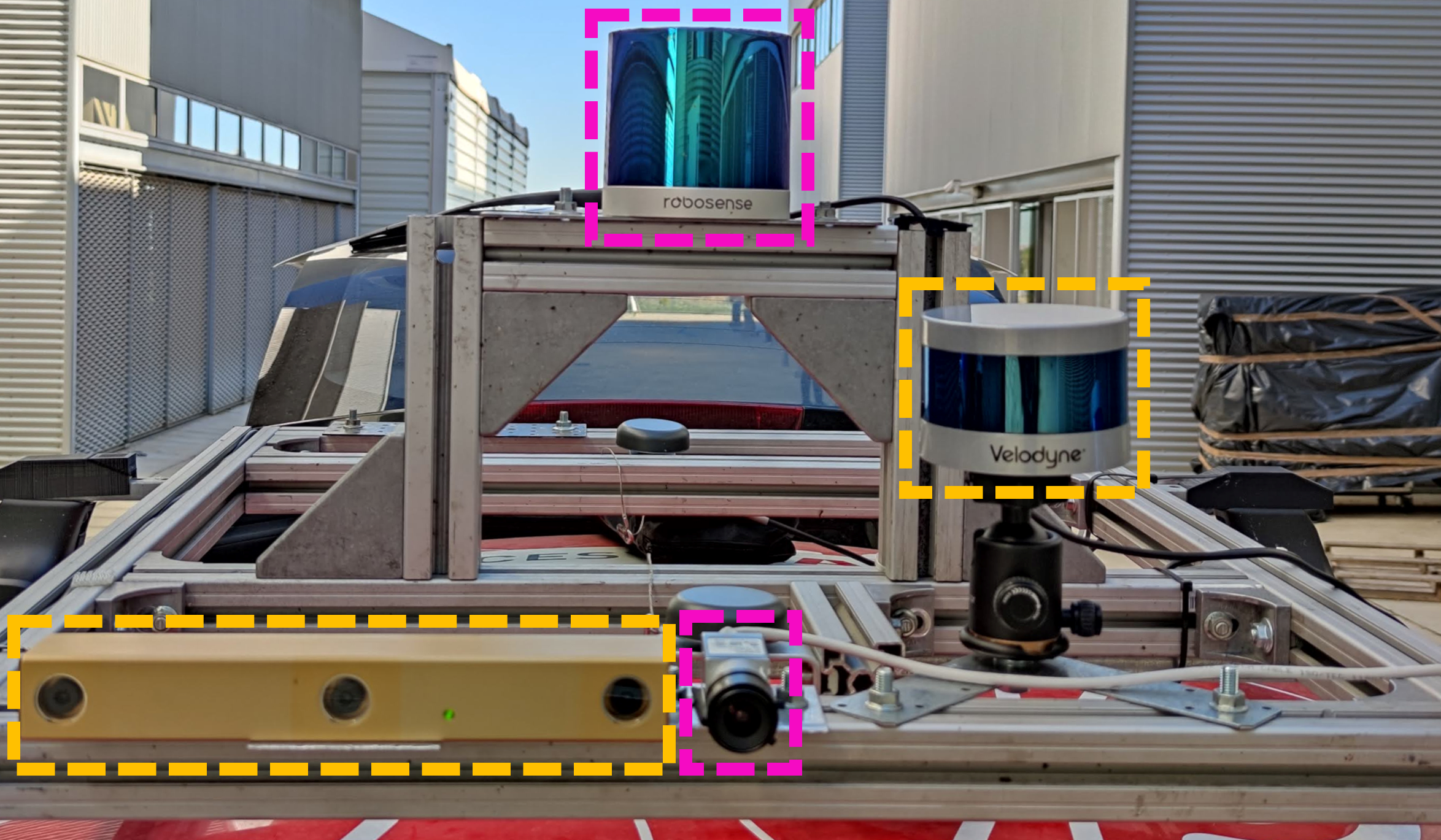}
        \label{fig:realsensors1b}
    }
\caption{The two sensor setups used in the real experiments. Calibrated pairs of devices are framed with the same color.}
\label{fig:realsensors1}
\end{figure}

For the first round, depicted in Fig. \ref{fig:realsensors1a}, two Velodyne VLP-16 LiDARs and a Bumblebee XB3 camera were mounted in a rig, with rotations emulating the ones that can be found in vehicle setups. 
In this step, we performed two different calibration procedures: monocular/LiDAR, involving one of the cameras of the stereo system and one of the LiDAR scanners, and LiDAR/LiDAR, between the two VLP-16 devices.

In the second round, we used the configuration shown in Fig. \ref{fig:realsensors1b}, with the Bumblebee XB3 stereo camera, a Basler 
acA2040-35gc camera with a 90\si{\degree} HFOV lens, a Robosense RS-LiDAR-32, and a Velodyne VLP-16 Hi-Res LiDAR. Here, we tested three different calibration alternatives: stereo/LiDAR, linking the XB3 and the VLP-16 Hi-Res scanner, monocular/LiDAR, this time with the wide-angle Basler camera and the RS-LiDAR-32, and monocular/monocular, between two of the cameras of the stereo system.

The sensors used in these experiments have very different features from each other; thus, the VLP-16 Hi-Res LiDAR has a tighter layer distribution than the regular VLP-16, whereas the RS-LiDAR-32 has twice as many scan planes, but they are irregularly spread, with much higher density around the central area. All the devices pose their own challenges for calibration, as the set of locations where the four circles of the calibration pattern are fully visible is much more limited than, for example, with the Velodyne HDL-64. As for the cameras, the narrow field of view exhibited by the XB3's cameras (43\si{\degree}) contrasts with the wide angle of the Basler. Overall, the number and variety of sensors and combinations used in the experiments ensure the generality of the results.
As with the synthetic experiments, points were extracted from the accumulation of $N=30$ frames, and $M=5$ target poses were used. The rest of the parameters remained unchanged from Table~\ref{tab:parameters}.

Ground truth of the relative position between sensors was not available, but some illustrative statistics about the performance of the calibration procedure with real sensors are presented below. On the one hand, Fig.~\ref{fig:ada_multipose} shows the dispersion of the estimated reference points across different poses of the calibration pattern, each represented by a point. Data from the five separate calibration procedures are included. The black line represents the mean, the dark shadow spans the standard deviation, and the light shadow covers 1.96 times the standard error of the mean.

\begin{figure}[htb]
\centering
\includegraphics[width=\linewidth]{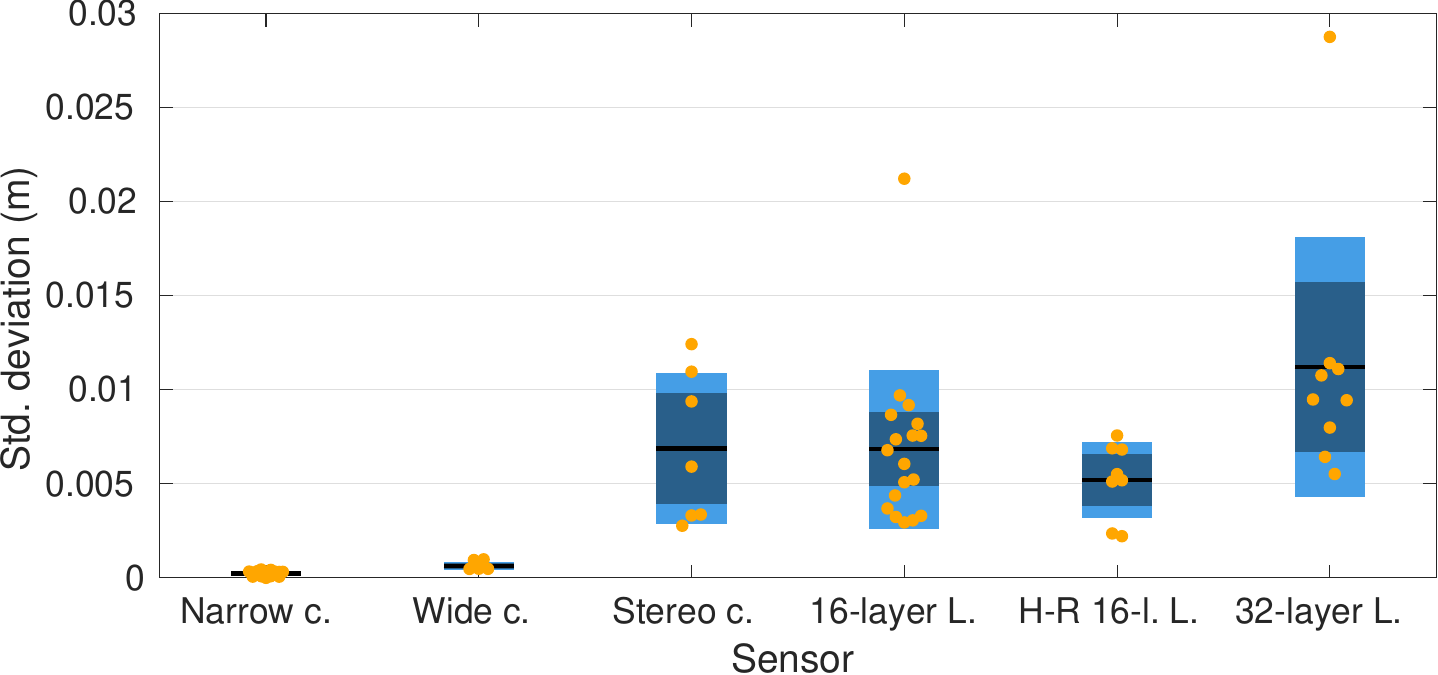}
\caption{Dispersion in the localization of the reference points in real experiments for the different camera (c.) and LiDAR (L.) devices.} 
\label{fig:ada_multipose}
\end{figure}
The results confirm that the dispersion in the LiDAR and stereo modalities is significantly higher than the one exhibited by its monocular counterpart, as suggested by the tests in the synthetic environment. However, the deviation is still small enough to enable higher accuracy in registration. It is possible to observe the presence of outliers corresponding to some particular poses of the calibration pattern; however, they do not raise relevant issues for the multi-pose calibration as they are well mitigated by the rest of the poses.

\begin{figure}[htb]
\centering
    \subfloat[]{%
        \includegraphics[width=0.48\linewidth]{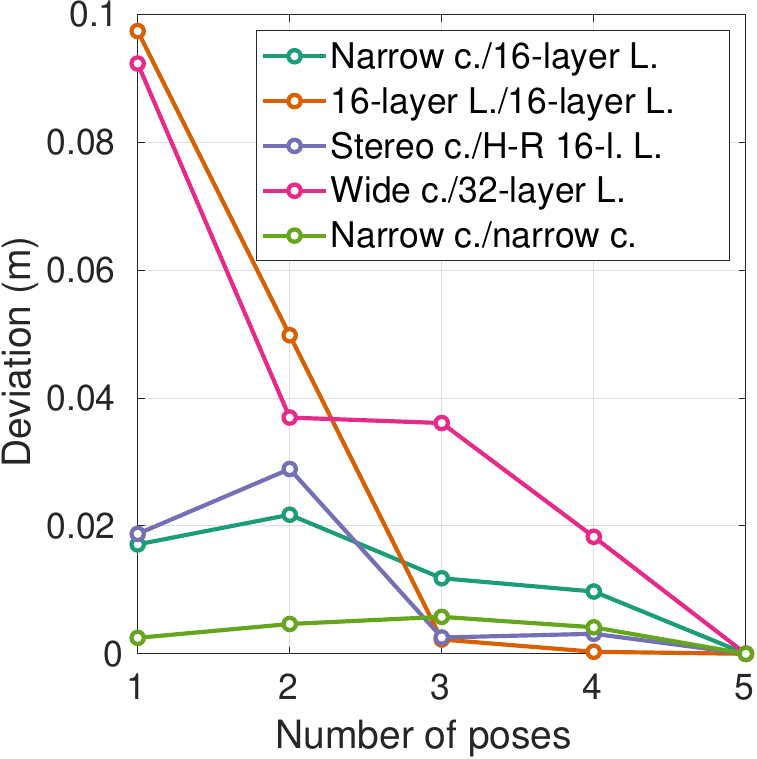}
    }
    \subfloat[]{%
        \includegraphics[width=0.48\linewidth]{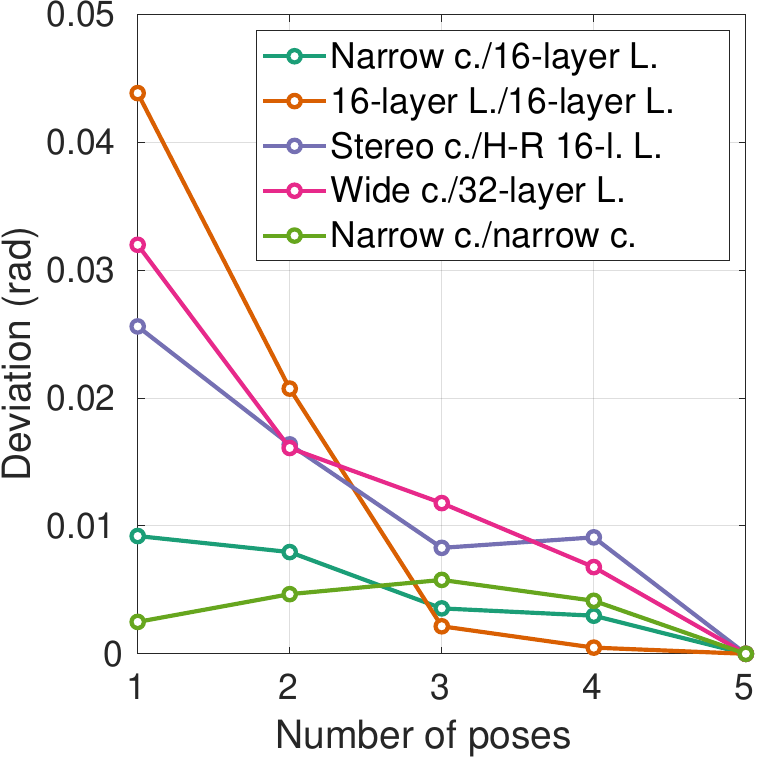}
    }
\caption{Linear and angular deviation from the final calibration result at $M=5$ in real experiments.}
\label{fig:adaposes2}
\end{figure}

On the other hand, Fig.~\ref{fig:adaposes2} shows the difference, measured in linear and angular errors, of the calibrations performed with $M \in [1, 4]$ versus the final result with $M=5$. The results validate the conclusion drawn in the previous section: using several pattern poses ($M>1$) causes significant changes in the calibration result up to 3 poses, where it plateaus. 

In the particular case of the narrow-angle camera/narrow-angle camera calibration, the result can be compared with the baseline provided by the manufacturer for the rectified stereo pair, yielding an average error across coordinates of \SI{2.73}{\milli\meter}.

\begin{figure*}[htb]
\centering
    \subfloat[]{%
        \includegraphics[width=0.49\linewidth]{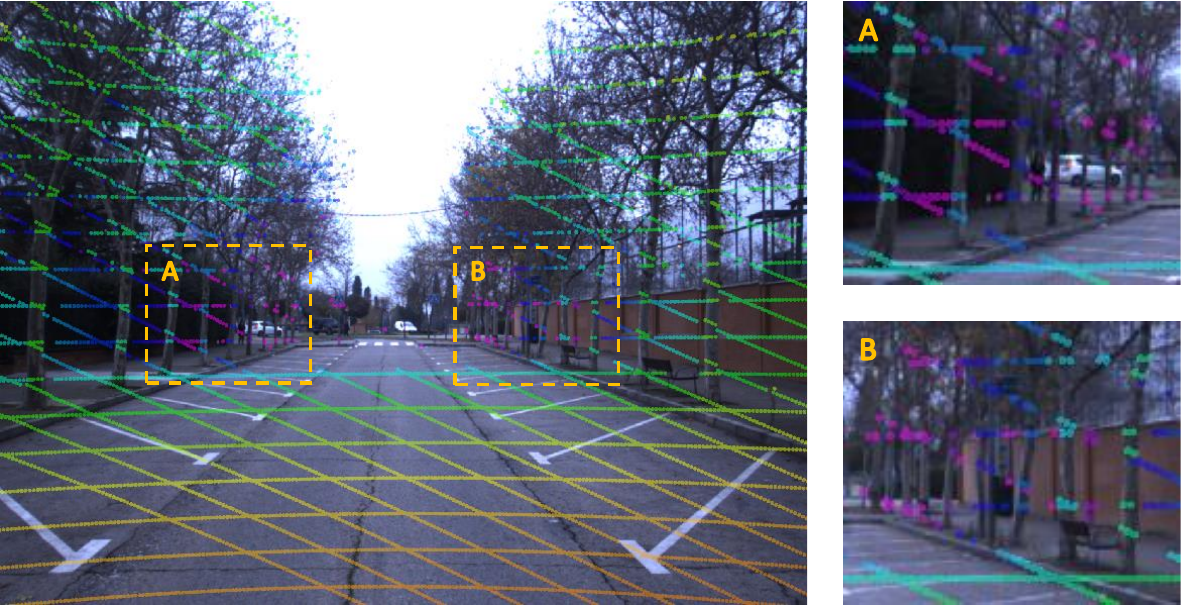}
        \label{fig:ada1}
    }
    \hfill
    \subfloat[]{%
        \includegraphics[width=0.49\linewidth]{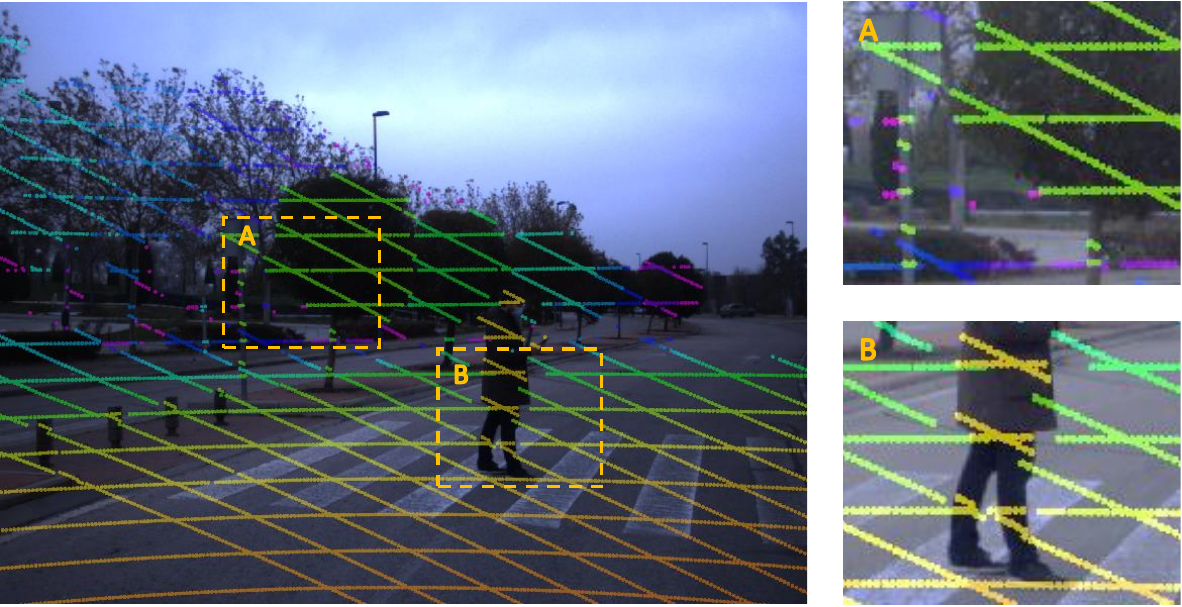}
        \label{fig:ada2}
    } 
    \\
    \subfloat[]{%
        \includegraphics[width=0.49\linewidth]{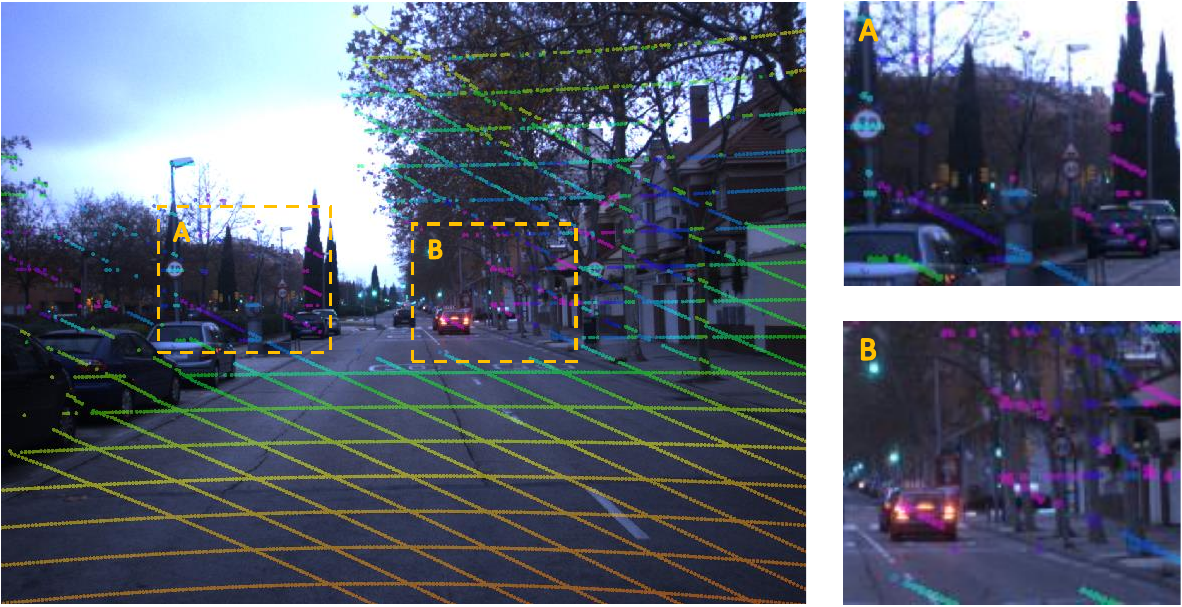}
        \label{fig:ada3}
    }
    \hfill 
    \subfloat[]{%
        \includegraphics[width=0.49\linewidth]{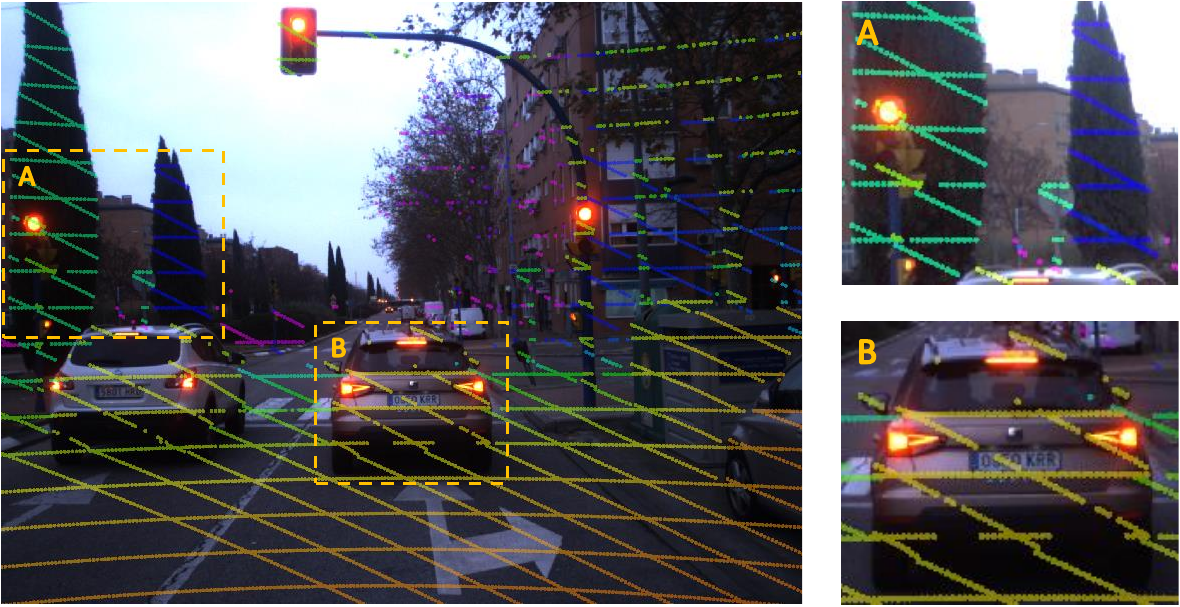}
        \label{fig:ada4}
    }
    \\
    \subfloat[]{%
        \includegraphics[width=0.49\linewidth]{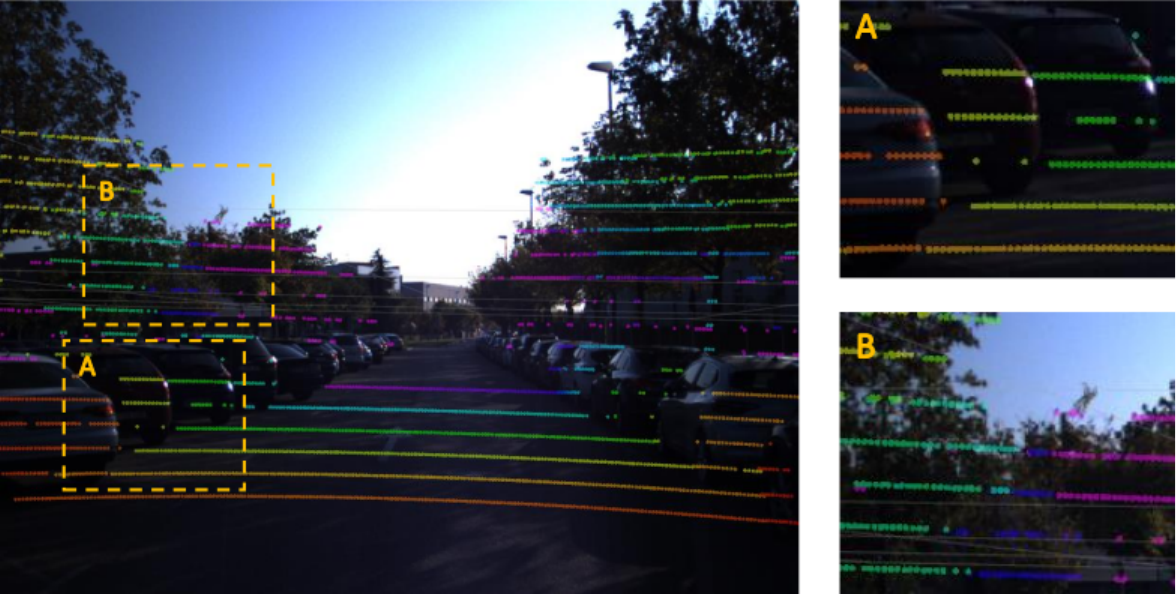}
        \label{fig:ada5}
    }
    \hfill
    \subfloat[]{%
        \includegraphics[width=0.49\linewidth]{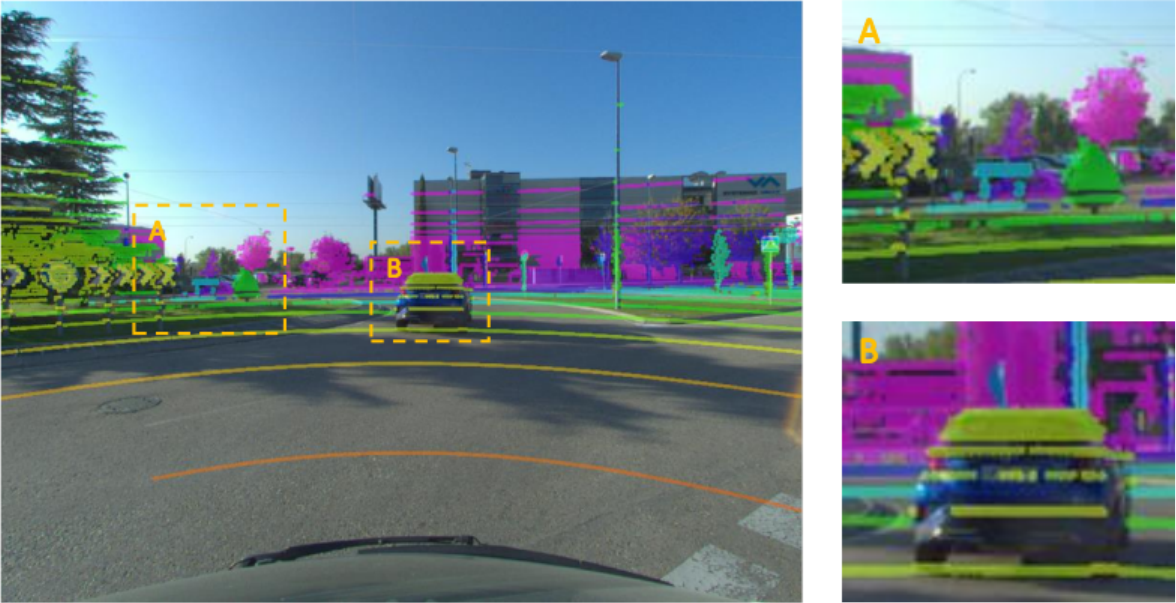}
        \label{fig:ada6}
    }
    \\
\caption{Samples (main view and two close-up views) of different traffic scenarios where LiDAR points have been projected onto the image using the sets of extrinsic parameters extracted with the proposed approach: narrow-angle camera/16-layer LiDAR and 16-layer LiDAR/16-layer LiDAR (a-d), stereo/Hi-Res 16-layer LiDAR (e), and wide-angle camera/32-layer LiDAR (f).}
\label{fig:ada_results}
\end{figure*}

Finally, Fig.~\ref{fig:ada_results} depicts various examples of traffic scenarios captured by the calibrated sensor setups, with specific regions enlarged so that the details can be well perceived. The first four correspond to the first sensor stack and illustrate the performance of the narrow-angle camera/16-layer LiDAR and 16-layer LiDAR/16-layer LiDAR calibrations. The last two show the results of the stereo/Hi-Res 16-layer LiDAR and wide-angle camera/32-layer LiDAR calibrations, respectively. As shown, the use of the extrinsic parameters extracted by the proposed approach enables a perfect alignment between both data modalities, even at a considerable distance from the car, being especially noticeable when representing thin objects (e.g., lamp poles or trees). 

\section{Conclusion}
\label{sec:conclusion}
We have presented an approach to obtain the extrinsic parameters representing the relative pose of any pair of sensors involving LiDARs, monocular or stereo cameras, of the same or different modalities. Unlike the existing works, the simplicity of the calibration scenarios and the characteristics provided by the proposed target allow obtaining accurate results for most sensing setups featured by autonomous vehicles. Moreover, minimal user intervention is required.

Additionally, we have introduced an advanced simulation suite that copes with the traditional imprecision at performance assessment and provides exact ground truth that enables a reliable evaluation of extrinsic calibration methods. 

Results obtained from the conducted experiments demonstrate that the algorithm presented in this work notably outperforms existing approaches. Tests performed over real data confirm the accuracy obtained in the simulation environment.

Nevertheless, the presented approach has room for improvement in certain aspects. Currently, manual pass-through filters are required to ease the target segmentation step in cluttered scenes. Introducing an automated target isolation process would remove the need for human intervention. On the other hand, as the relative pose between the target and the sensors has an influence on the accuracy of the reference point extraction, developing a guided method that guarantees sufficient variability of the different target locations during the multi-pose approach, now selected by the human operator, would likely enhance the quality of the calibration result.

Some other complementary lines of work remain open for the future. An outlier rejection scheme might be useful to discard spurious samples obtained in the reference point extraction procedure. At this point, accurate modeling of the sensor noise could be convenient, which will also enable adapting the parameter settings to each particular device. Besides, the proposed method has been designed to determine a fixed set of extrinsic parameters before the perception system is deployed; however, sensor setups mounted in movable platforms, such as autonomous vehicles, can suffer miscalibrations during regular operation. The use of the proposed method would require the ability to detect these situations early, prompting the user to perform a recalibration when necessary.  

Although there is still a road ahead, this proposal provides a practical approach to solve a common problem for the scientific community working in this field, bringing autonomous driving and robotics solutions closer to their final deployment. 


\ifCLASSOPTIONcaptionsoff
  \newpage
\fi



%
\bibliographystyle{IEEEtran}
\bibliography{paper}

%

\begin{IEEEbiography}[{\includegraphics[width=1in,height=1.25in,clip,keepaspectratio]{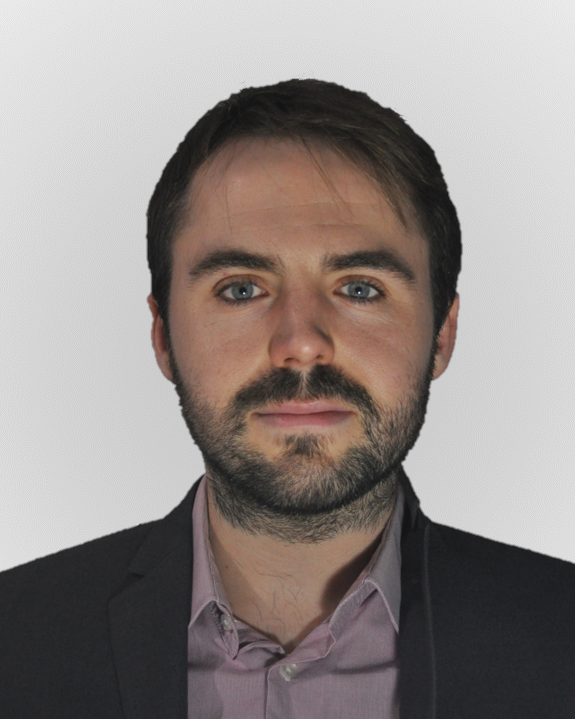}}]{Jorge Beltrán} received his M.Sc degree in Robotics and Automation from Universidad Carlos III de Madrid, Spain, in 2016. He is currently pursuing his Ph.D. in Electrical, Electronic and Automatic Engineering at the same University. His research work lies in the field of perception systems for autonomous vehicles, focusing on multi-modal sensor calibration, sensor fusion, and 3D object detection using deep neural networks.
\end{IEEEbiography}

\begin{IEEEbiography}[{\includegraphics[width=1in,height=1.25in,clip,keepaspectratio]{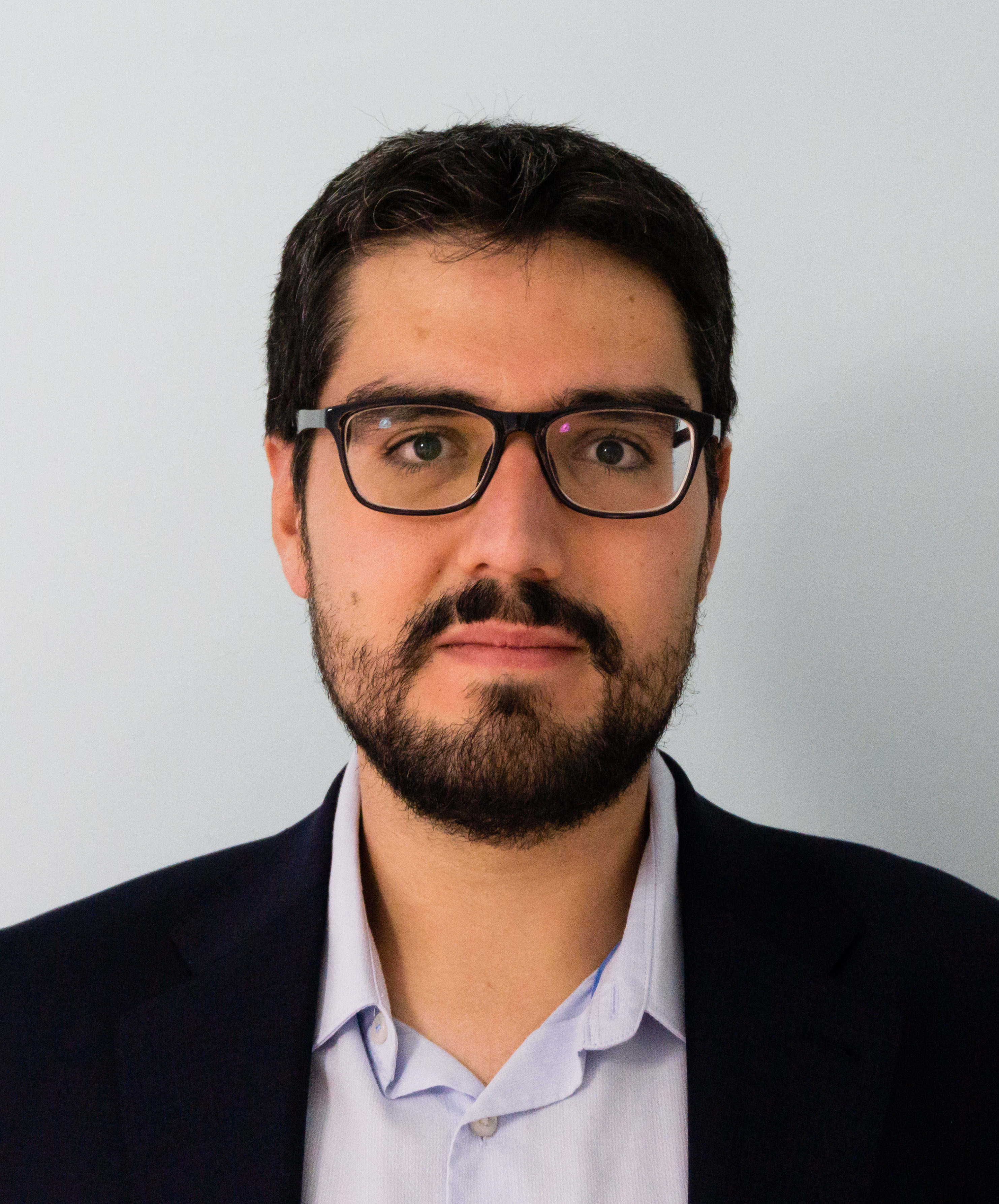}}]{Carlos Guindel} received the Ph.D. degree in Electrical Engineering, Electronics and Automation from the University Carlos III of Madrid, Spain, in 2019, where he is currently a postdoctoral researcher. His research focuses on the application of deep learning techniques to the intelligent transportation systems field, covering topics such as object detection, pose estimation, and sensor fusion.
\end{IEEEbiography}

\begin{IEEEbiography}[{\includegraphics[width=1in,height=1.25in,clip,keepaspectratio]{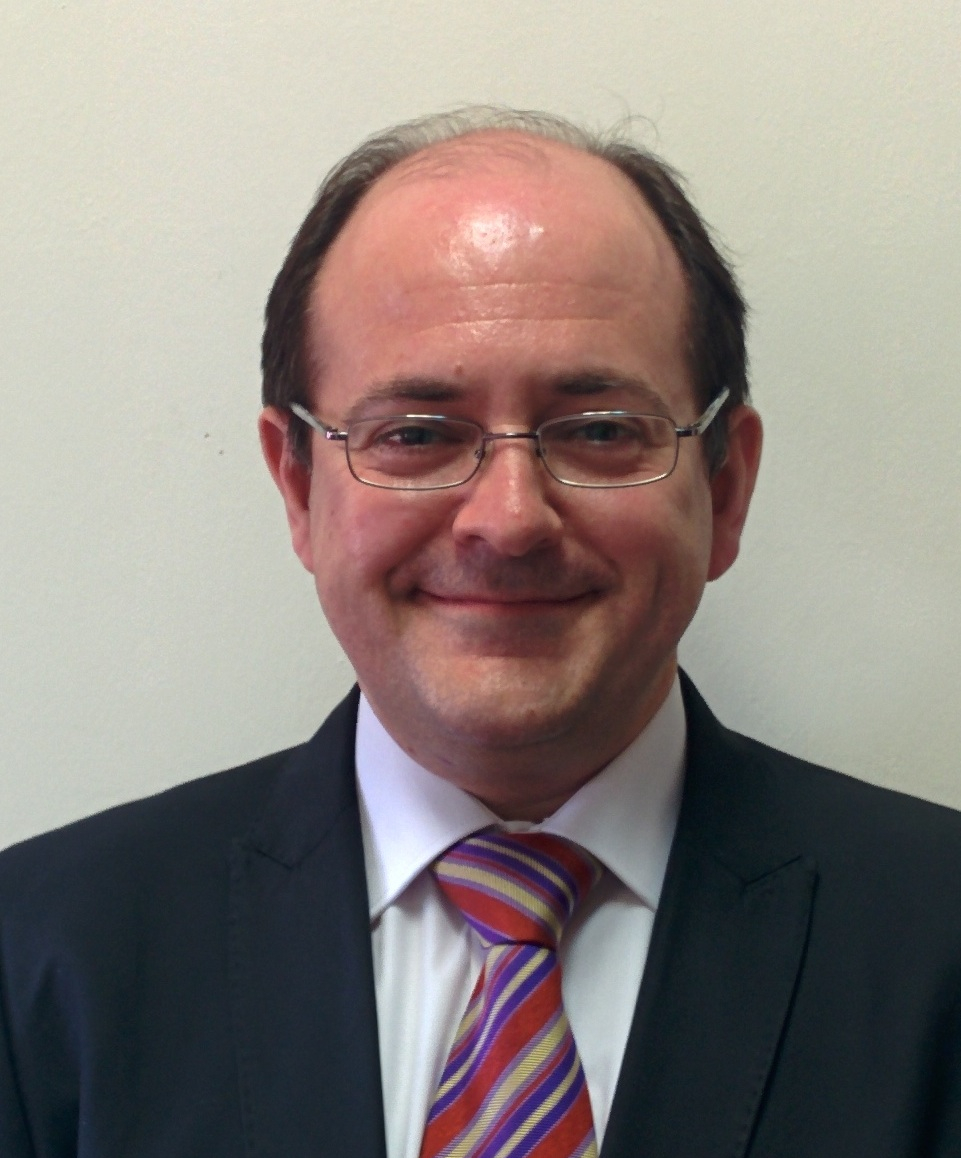}}]{Arturo de la Escalera}  obtained his Ph.D. degree in Robotics in 1995 from Universidad Politecnica de Madrid (Spain).  In 1993, he joined the Department of Systems Engineering and Automation at Universidad Carlos III de Madrid (Spain), where he became an Associate Professor in 1997 and Full Professor in 2018.  His current research interests include Robotics and Intelligent Transportation Systems, with special emphasis on environment perception.
\end{IEEEbiography}

\begin{IEEEbiography}[{\includegraphics[width=1in,height=1.25in,clip,keepaspectratio]{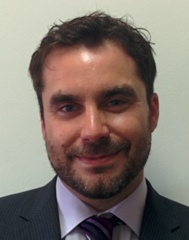}}]{Fernando Garc\'{i}a} received his Ph.D. degree in Electrical,  Electronic and  Automatic Engineering from Universidad Carlos III de Madrid in 2012 where he works as Associate Professor. His research interests are perception and data fusion,  mainly applied to vehicles and robotics. He is member of the Board of governors of the IEEE-ITS Society since 2017 and chair of the  Spanish chapter for the period  2019-2020. He 
was recipient of IEEE ITS Young Researcher/Engineer Award 2021.
\end{IEEEbiography}


\vfill


\end{document}